\pdfoutput=1
\documentclass[11pt]{article}

\usepackage[]{acl}

\usepackage{times}
\usepackage{latexsym}

\usepackage{amsmath, amssymb, amscd, amsthm, amsfonts}
\usepackage{mathtools,enumerate,mathrsfs,graphicx}
\usepackage{subfigure}
\usepackage{multirow}
\usepackage{hyperref}
\usepackage{algorithm}
\usepackage{algorithmicx}
\usepackage[inline]{enumitem}
\usepackage{algpseudocode}
\usepackage{stfloats}
\usepackage{tcolorbox}
\usepackage{booktabs}
\usepackage{amsmath}
\usepackage{float}

\usepackage[T1]{fontenc}
\usepackage[utf8]{inputenc}

\usepackage{microtype}

\usepackage{sistyle}
\SIthousandsep{,}

\title{ROSE: Robust Selective Fine-tuning for Pre-trained Language Models}

\author{
Lan Jiang$^{1}$\thanks{\ \ Equal contribution}, Hao Zhou$^{2}$\footnotemark[1], Yankai Lin$^{3,4}$\thanks{\ \ Part of the work was done while Yankai Lin and Peng Li were working at Tencent.}, Peng Li$^{5}$\footnotemark[2], \textbf{Jie Zhou$^{2}$}, \textbf{Rui Jiang$^{1}$\thanks{\ \ Corresponding Author.}}  \\
$^1$Ministry of Education Key Laboratory of Bioinformatics, Center for Synthetic and Systems Biology, \\
Department of Automation, BNRist, Tsinghua University, China. \\
$^2$Pattern Recognition Center, WeChat AI, Tencent Inc., China \\
$^3$Gaoling School of Artificial Intelligence, Renmin University of China, Beijing, China \\
$^4$Beijing Key Laboratory of Big Data Management and Analysis Methods , Beijing, China \\
$^5$Institute for AI Industry Research (AIR), Tsinghua University, China. \\
\texttt{jiangl20@mails.tsinghua.edu} \\
}

\begin{document}
\maketitle

\begin{abstract}
Even though the large-scale language models have achieved excellent performances, they suffer from various adversarial attacks.
A large body of defense methods has been proposed. 
However, they are still limited due to redundant attack search spaces and the inability to defend against various types of attacks.
In this work, we present a novel fine-tuning approach called \textbf{RO}bust \textbf{SE}letive fine-tuning (\textbf{ROSE}) to address this issue.
ROSE conducts selective updates when adapting pre-trained models to downstream tasks, filtering out invaluable and unrobust updates of parameters.
Specifically, we propose two strategies: the first-order and second-order ROSE for selecting target robust parameters.
The experimental results show that ROSE achieves significant improvements in adversarial robustness on various downstream NLP tasks, and the ensemble method even surpasses both variants above.
Furthermore, ROSE can be easily incorporated into existing fine-tuning methods to improve their adversarial robustness further.
The empirical analysis confirms that ROSE eliminates unrobust spurious updates during fine-tuning, leading to solutions corresponding to flatter and wider optima than the conventional method.
Code is available at \url{https://github.com/jiangllan/ROSE}.
\end{abstract}
\section{Introduction}

\begin{figure*}[htb]
    \centering
    \subfigure[w/o Fine-Tuning]{
		\begin{minipage}[b]{0.23\textwidth}
			\includegraphics[width=1\textwidth]{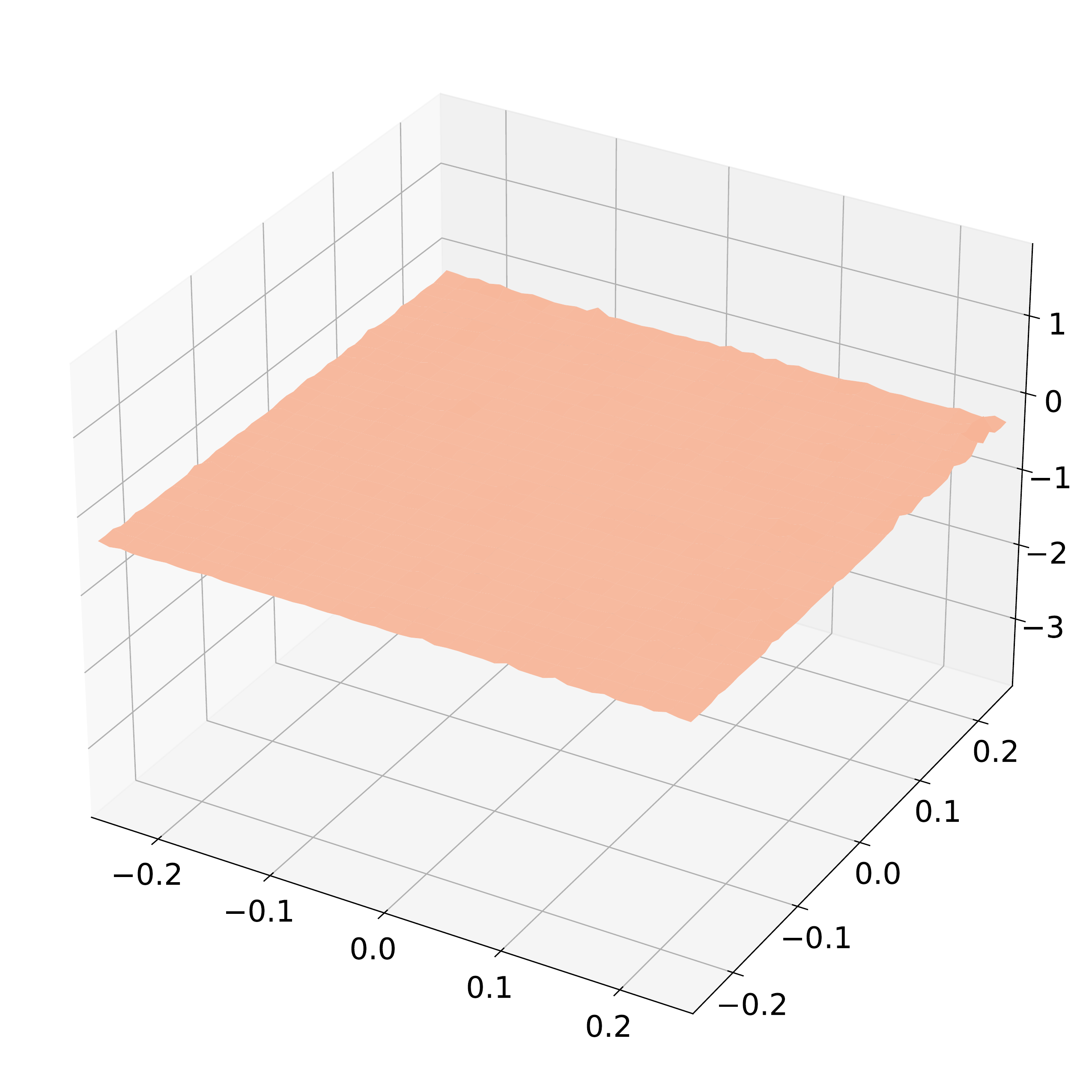}
		\end{minipage}
	}
	\subfigure[Overfit]{
		\begin{minipage}[b]{0.23\textwidth}
			\includegraphics[width=1\textwidth]{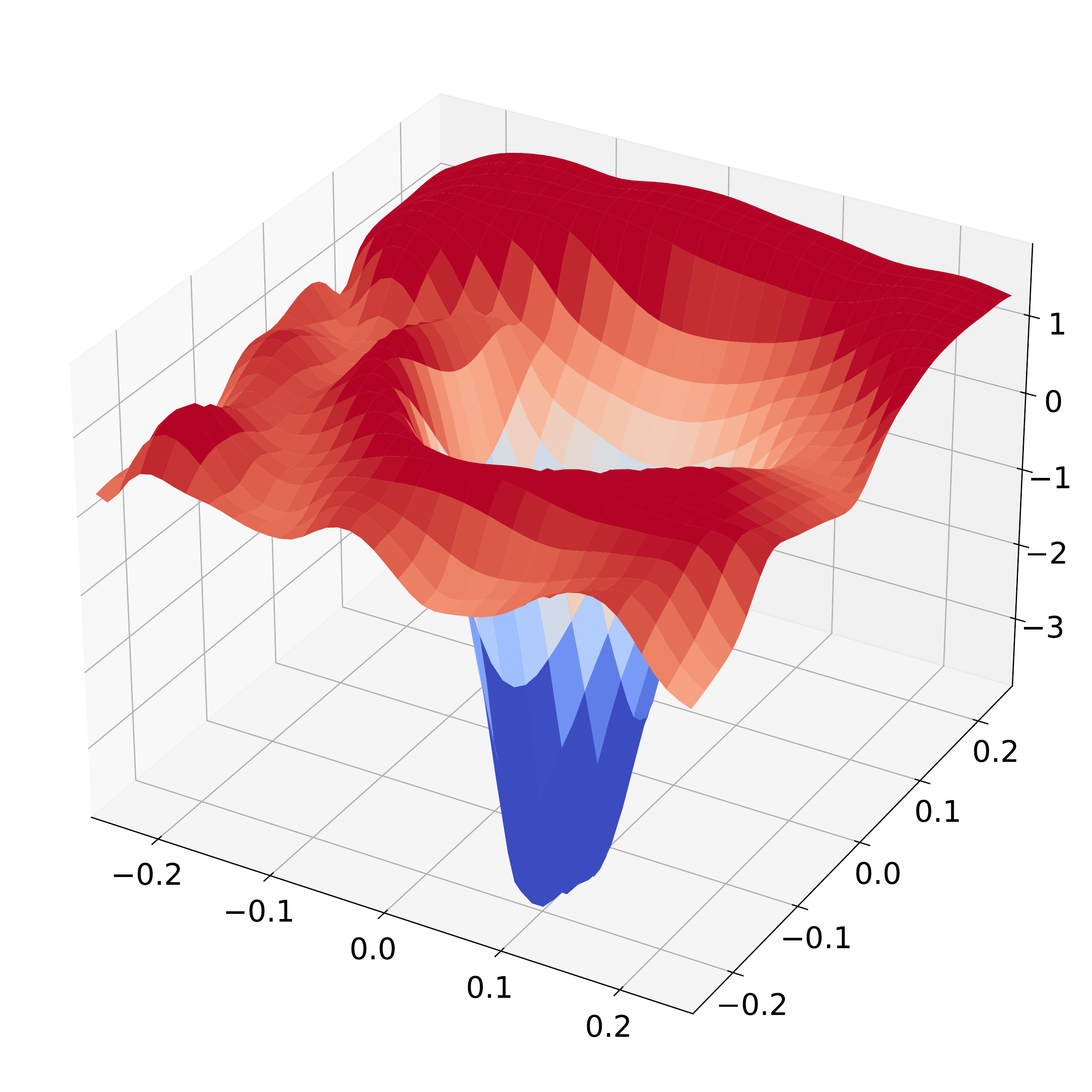}
		\end{minipage}
% 		\label{fig:ll-1D-vanilla}
	}\hspace{2.25mm}
	\subfigure[Vanilla Fine-Tuning]{
		\begin{minipage}[b]{0.23\textwidth}
			\includegraphics[width=1\textwidth]{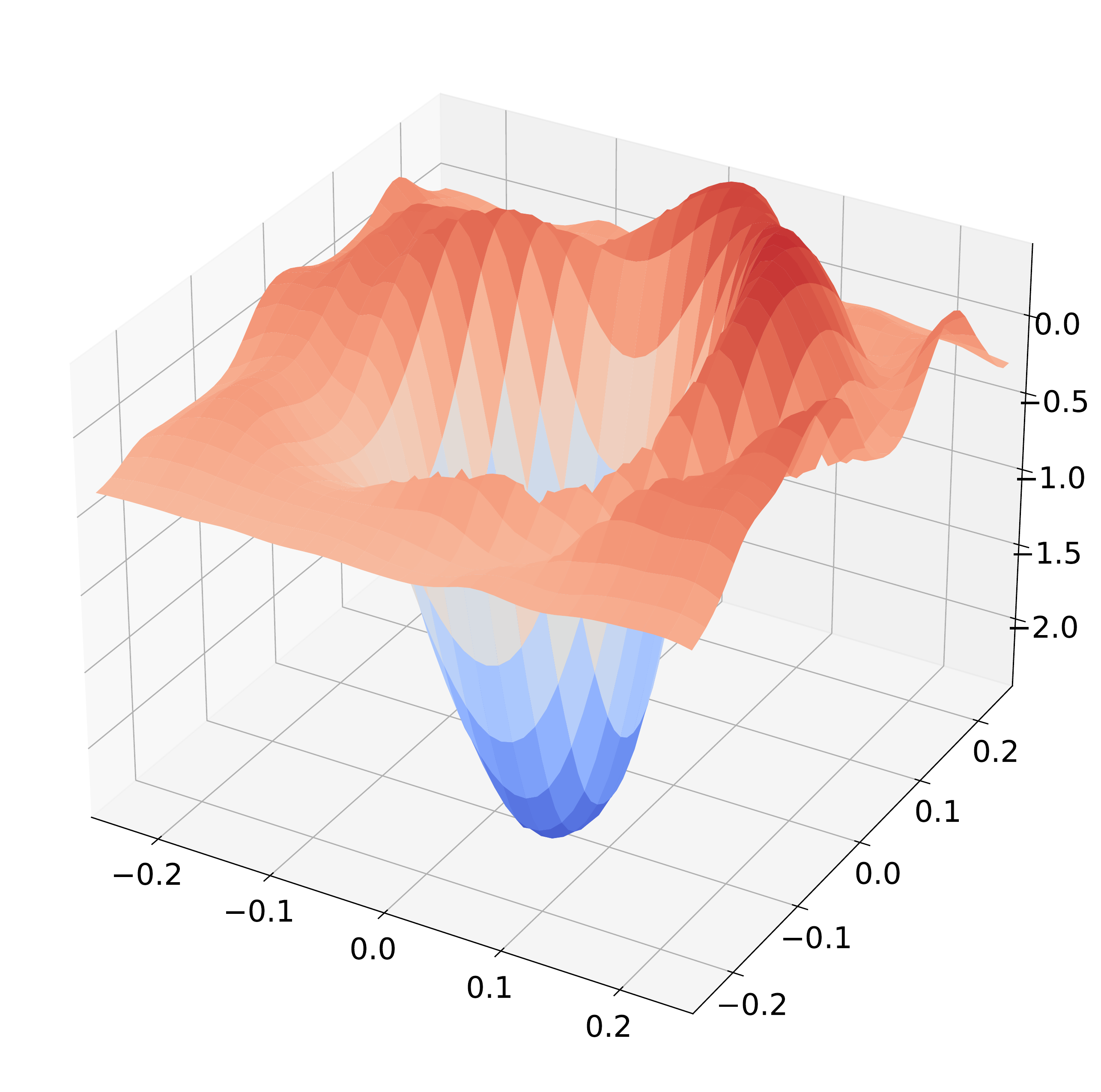}
		\end{minipage}
% 		\label{fig:ll-1D-vanilla}
	}
    \subfigure[ROSE-Ensemble]{
    	\begin{minipage}[b]{0.23\textwidth}
   		\includegraphics[width=1\textwidth]{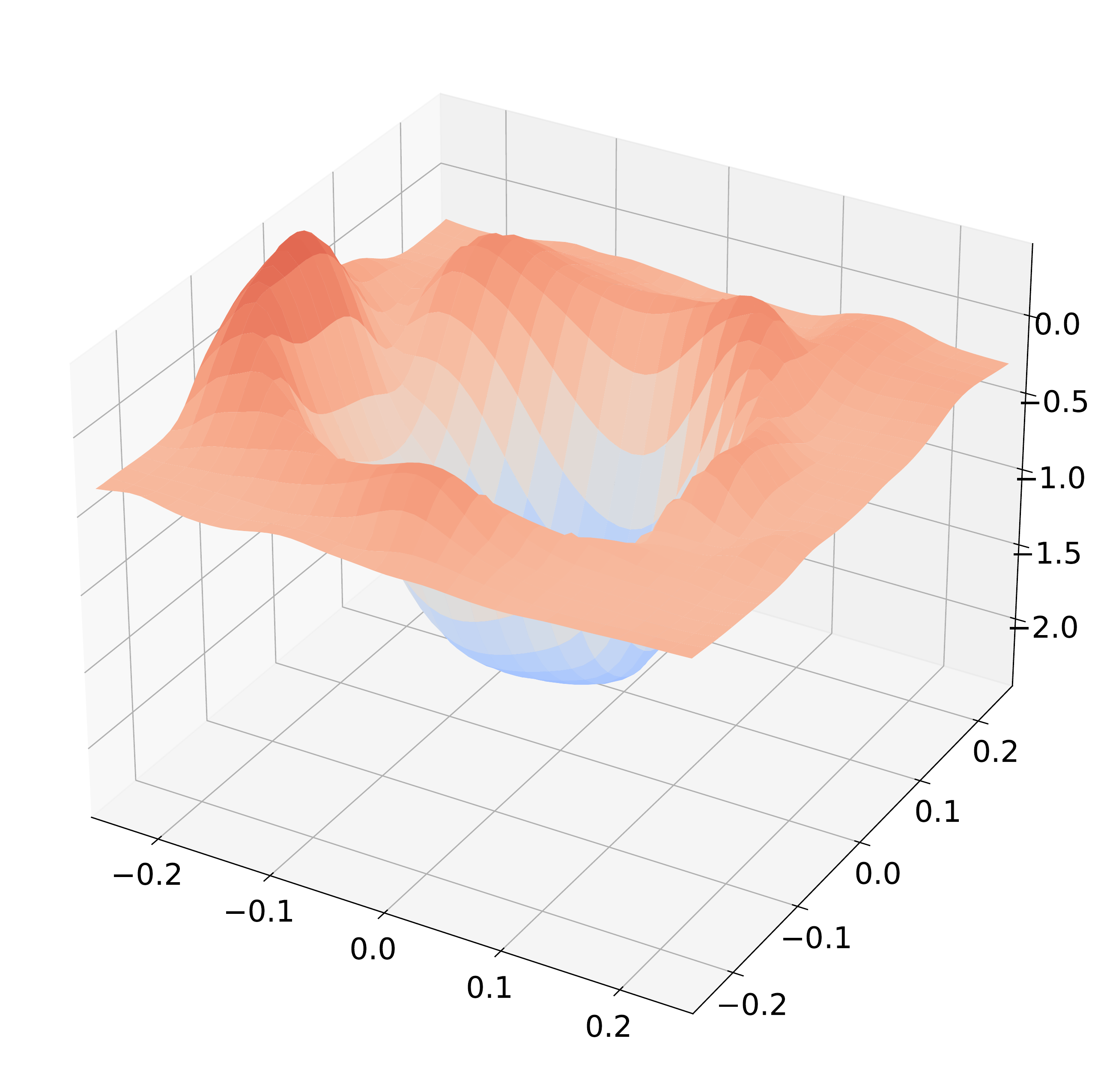}
    	\end{minipage}
% 	\label{fig:ll-1D-first}
    }
    \caption{The loss surfaces of fine-tuned models with different fine-tuning strategies, initialized from RoBERTa$\rm_{\texttt{BASE}}$. 
    ROSE leads to wider and flatter optima, enhancing the adversarial robustness of adapted models.}
    \label{fig:cherry-picking}
\end{figure*}

Fine-tuning large-scale pre-trained language models has become a dominant paradigm recently, achieving remarkable performances across various natural language processing benchmarks.
However, recent studies \cite{ribeiro-etal-2020-beyond,Jin_Jin_Zhou_Szolovits_2020,nie-etal-2020-adversarial,lin-etal-2021-using,jiang-etal-2022-length} have highlighted the lack of adversarial robustness in models fine-tuned on specific downstream tasks, \textit{i}.\textit{e}., adapted models are vulnerable to various types of adversarial attacks.
Adversarial examples mostly fool models via character or word-level perturbations which are either those tokens not appearing in training sets or superficial cues associated with labels~\cite{DBLP:journals/corr/abs-2104-00312,le-etal-2022-shield}.
The vulnerability of adapted models can be attributed to their tendency to capture shallow and spurious patterns when fine-tuning on downstream tasks, instead of utilizing the general linguistic knowledge they have learned during the pre-training stage~\cite{pmlr-v119-sagawa20a,warstadt-etal-2020-learning,DBLP:conf/iclr/GoukHP21,DBLP:conf/nips/DongLLYZ21}.

To address this issue, various defense methods have been proposed, including adversarial training~\cite{DBLP:journals/corr/GoodfellowSS14,DBLP:conf/iclr/ZhuCGSGL20,ivgi-berant-2021-achieving}, adversarial data augmentation \cite{zhang-etal-2019-paws,zheng-etal-2020-evaluating,si-etal-2021-better} and so on.
Adversarial training and adversarial data augmentation stand to provide the most promising performance among all the defense methods.
They enhance adversarial robustness by re-training models with additional adversarial data generated either via human-crafting or by conducting projected gradient ascent on the benign examples.
In essence, these methods prevent models from learning misleading features by covering more diverse training data.
However, they are limited to practice as they often require prohibitively large attack search spaces, and are not generally applicable to different types of attacks.
% We try to address these challenges from a learning perspective, which enables models to clarify robust features and perform selective updates actively.

In this work, we present an attack-agnostic and model-agnostic defense method called \textbf{RO}bust \textbf{SE}lective Fine-Tuning (\textbf{ROSE}) to address these challenges from a learning perspective.
ROSE is a novel fine-tuning method that conducts robust updates selectively during the fine-tuning stage.
The intuition behind our method is straightforward: \textit{only robust and informative updates of parameters should be conducted}.
While the improper ones, which make the fine-tuned model capture superficial cues and overfit the training data of the downstream task, should be filtered out.

Specifically, we propose two strategies in response to the above challenges: first-order ROSE and second-order ROSE.
Our first-order ROSE employs adversarial perturbations to clarify parameters that are robust against slight perturbation in the hidden space, enabling models to cope with superficial patterns among examples with similar semantics in the current step. 
% Inspired by recent works \cite{gao-etal-2021-simcse,NEURIPS2021_5a66b920} on \textit{Dropout}, we employ Dropout to generate perturbed inputs with little overhead costs.
Our second-order ROSE allows models to counter superficial patterns across examples with different semantics along the fine-tuning process by smoothing the optimization trajectory.
% It is done by smoothing the trajectory of the conventional optimizer.
% The corresponding second-order robustness measures the smoothness of the current updates compared to former updates and filters out parameters with aggressive updates.
We also propose an ensemble method to aggregate the benefits of the above two strategies.
ROSE distinguishes parameters based on robust criteria at each step in the backward process, then performs robust updates while freezing the remaining parameters.

Figure \ref{fig:cherry-picking} illustrates the loss landscapes of solutions found by different fine-tuning strategies (pre-trained initial, vanilla fine-tuning, overfitting and ROSE-tuning) on specific tasks.
ROSE leads to solutions corresponding to broader and flatter optima compared to traditional fine-tuning methods, which implies that it achieves better adversarial robustness as found in~\newcite{DBLP:journals/corr/GoodfellowV14}.
Moreover, our probing experiment illustrates that ROSE prefers deeper linguistic features rather than shallow lexical ones during fine-tuning.
The above empirical analysis confirms the inner working of ROSE.
ROSE allows for more robust solutions by masking out unreliable and spurious updates when fine-tuning models on downstream tasks.

We conduct extensive experiments to evaluate the effectiveness of ROSE.
We compare ROSE to several strong defense methods.
The results show that ROSE exhibits superior adversarial robustness on challenging examples, and achieves comparable or even better benign performance on several benchmarks.
ROSE is generic and can be incorporated with existing methods to further enhance their adversarial robustness.

% We summarize our contributions as follow:
% \begin{enumerate}[wide, labelwidth=!, topsep=1pt, itemsep=1pt, partopsep=1pt, labelindent=10pt]
%     \item We propose a novel robust fine-tuning method ROSE with three variants. 
%     ROSE is attack-agnostic and can be easily integrating into existing methods to further improve their robustness.
%     \item Experimental results show that ROSE achieves state-of-the-art performance on several popular adversarial benchmarks. 
%     Empirical analysis demonstrate ROSE finds solutions with wider and flatter loss surfaces.
% \end{enumerate}

\section{Methodology}

In this section, we will introduce our method in detail.
The key motivation of our method is to select parameters that carry stable information for downstream tasks during fine-tuning.
Specifically, the vanilla fine-tuning updates all the parameters in the backward process, while ROSE only updates the robust and informative ones.
To identify robust parameters, we propose two robustness criteria and corresponding selective fine-tuning strategies: first-order ROSE (Section \ref{sec:sparse-first}) and second-order ROSE (Section \ref{sec:sparse-second}).
% To identify robust parameters, we propose two robustness criteria and corresponding selective fine-tuning strategies: first-order ROSE which selects parameters with first-order robustness (Section \ref{sec:sparse-first}), and second-order ROSE which selects second-order robust ones (Section \ref{sec:sparse-second}).
Furthermore, we explore an ensemble robust selective fine-tuning method (Section \ref{sec:sparse-ensemble}), which aggregates the benefits of the above two strategies.
The overall training algorithm of ROSE when applied to AdamW \cite{DBLP:conf/iclr/LoshchilovH19} is shown in Algorithm \ref{alg:sparse}.

\begin{algorithm*}[ht]
    \renewcommand{\algorithmicrequire}{\textbf{given}}
	\renewcommand{\algorithmicensure}{\textbf{initialize}}
	\renewcommand{\algorithmicreturn}{\textbf{return}}
	\renewcommand{\algorithmicrepeat}{\textbf{repeat}}
    \caption{ROSE for AdamW Optimizer}
    \label{alg:sparse}
    \begin{algorithmic}[1]
        \Require $\alpha=0.001,\beta_1,\beta_2\in[0,1),\epsilon=10^{-8},\lambda\in\mathbb{R}$
        \State \textbf{initialize} time step $t\gets 0$, parameter vector $\boldsymbol{\theta}_{t=0}\in\mathbb{R}^n$, first moment vector $\mathbf{m}_{t=0}\gets \mathbf{0}$, second moment vector $\mathbf{v}_{t=0}\gets \mathbf{0}$, learning rate $\eta\in\mathbb{R}$
        \Repeat
        \State $t \gets t+1$
        \State $\mathbf{g}_t \gets \nabla  \mathcal{L}_t^{\text{SCE}}(\boldsymbol{\theta}_{t-1})$
        \State $\mathbf{r}_t^{\text{F}}\gets\Vert\nabla \mathcal{L}_t^{\text{KL}}(\boldsymbol{\theta}_{t-1})\Vert_F$ \Comment{calculate first-order risks}
        \State $\mathbf{r}_t^{\text{S}}\gets\lvert(1-\beta_1)\Vert\mathbf{g}_t\Vert_F / \Vert\mathbf{m}_{t-1}\Vert_F-1\rvert$ \Comment{calculate second-order risks}
        \State $\mathbf{M}_t\gets\text{CalculateMask}(\mathbf{r}_t^{\text{F}}, \mathbf{r}_t^{\text{S}})$ \Comment{calculate mask}
        \State $\mathbf{g}_t^{'}\gets\mathbf{M}_t\odot\mathbf{g}_t+(\mathcal{J}-\mathbf{M}_t)\odot\mathbf{m}_{t-1}$ \Comment{update gradients}
        \State $\mathbf{m}_t \gets \beta_1 \mathbf{m}_{t-1} + (1-\beta_1)\mathbf{g}_t^{'}$
        \State $\mathbf{v}_t \gets \beta_2 \mathbf{v}_{t-1} + (1-\beta_2)\mathbf{{g}_t^{'}}^2$
        \State $\hat{\mathbf{m}_t}\gets\mathbf{m}_t/(1-\beta_1^t)$
        \State $\hat{\mathbf{v}_t}\gets\mathbf{v}_t/(1-\beta_2^t)$
        \State $\boldsymbol{\theta}_t\gets\boldsymbol{\theta}_{t-1}-\eta\bigg(\hat{\mathbf{m}_t}/(\sqrt{\hat{\mathbf{v}_t}}+\epsilon)+\lambda\boldsymbol{\theta}_{t-1} \bigg)\odot\mathbf{M}_t$ \Comment{update weights}
        \Until \textit{stopping criterion is met}\\
        \Return optimized parameters $\boldsymbol{\theta}_t$
    \end{algorithmic}
\end{algorithm*}

\subsection{First-order ROSE}
\label{sec:sparse-first}
Our first-order ROSE aims to select parameters that are insensitive to first-order perturbation in the hidden space.
We employ adversarial inputs to distinguish robust parameters.
Different from the conventional virtual adversarial examples generated via PDG-based methods, ROSE adopts dropout to generate adversarial perturbation with little overhead cost.
We follow the method used in \newcite{gao-etal-2021-simcse,NEURIPS2021_5a66b920}, which passes the same input to the model twice in the forward process with different dropout, and obtains two outputs correspondingly.
Then in the backward process, ROSE only updates parameters that are insensitive to the difference between the two outputs.

Formally, we denote an initial pre-trained model as $\boldsymbol{\theta}_0$, the two probability distributions produced with different dropout at the $t$-step as $\mathcal{P}_t, \mathcal{P}_t^{'}$, and the Kullback-Leibler (KL) divergence between them is defined as follows:
\begin{eqnarray}
\label{eq:1}
    \mathcal{L}^\text{KL}_t = \mathcal{D}_{KL}(\mathcal{P}_t\Vert\mathcal{P}^{'}_t) + \mathcal{D}_{KL}(\mathcal{P}_t^{'}\Vert\mathcal{P}_t).
\end{eqnarray}

In the backward process, first-order ROSE filters out parameters which incline to learn superficial cues between similar examples.
The potential risk $\mathit{r}_t^{\text{F},i}$ of the $i$-th parameter in model is computed as the $\ell_F$ norm of the gradient with regard to $\mathcal{L}^\text{KL}_t$:
\begin{eqnarray}
\label{eq:2}
    \mathit{r}_t^{\text{F},i} = \Vert(\nabla \mathcal{L}^\text{KL}_t)_i(\boldsymbol{\theta}_{t-1})\Vert_F.
\end{eqnarray}
Then we sort the magnitude of the sensitivity from $\mathit{r}_t^{\text{F},1},  \mathit{r}_t^{\text{F},2}, \cdots, \mathit{r}_t^{\text{F},n}$ into $\mathit{r}_t^{\text{F},k_1}, \mathit{r}_t^{\text{F},k_2}, \cdots, \mathit{r}_t^{\text{F},k_n}$ in ascending order.
% First-order ROSE filters out the parameters that are relatively fragile to perturbation in first-order hidden space, which is $\mathcal{L}^\text{KL}_t$ there.
% In this way, First-order ROSE prevents models from learning superficial cues on similar examples.

Given the upper threshold $c_h^F$ (\textit{e.g.}, $60\%$) for robust parameters, the mask $\mathbf{M}_t^{\text{F}}$ can be derived as:
\begin{eqnarray}
\label{eq:3}
    \mathbf{M}_t^{\text{F},k_i} = 
    \begin{cases}
    1 & i/n \leq c_h,\\
    0 & \text{otherwise.}
    \end{cases}
\end{eqnarray}

Note that, only the classification loss will be used to update the weights of models, while gradients with regard to $\mathcal{L}^\text{KL}_t$ will be discarded after calculating masks.
% In other words, we set the gradients of optimizer to zero after calculating the mask.

\subsection{Second-order ROSE}
\label{sec:sparse-second}
Our second-order ROSE smooths the optimization trajectory to prevent models from learning spurious patterns between different groups of data points along the fine-tuning process.
More precisely speaking, our second-order ROSE selects and tunes parameters that are less aggressively updated to avoid overfitting on spurious patterns.
A straightforward idea is to calculate the second derivatives of the classification loss as the second-order risks.
Unfortunately, it requires prohibitive computation and storage costs.
Thus we employ a stochastic gradient-based optimizer like AdamW to approximate this solution.

Formally, we denote the softmax cross-entropy loss at the $t$-step as $\mathcal{L}_t^\text{SCE}$, and the first momentum of optimizer as $\mathbf{m}_{t-1}$.
Then the second-order risk $\mathit{r}_t^{\text{S},i}$ of the $i$-th parameter in the model is defined as the relative magnitude between current gradients $\mathbf{g}_t$ and the exponential moving average $\mathbf{m}_{t-1}$, which is computed as:
\begin{eqnarray}
\label{eq:4}
     \mathit{r}_t^{\text{S},i} = \bigg\lvert \frac{\alpha\Vert \mathbf{g}_t^i\Vert_F}{\Vert\mathbf{m}_{t-1}^i\Vert_F} - 1\bigg\rvert,
\end{eqnarray}
where $\alpha$ is a scaling coefficient.
In AdamW, $\alpha=(1-\beta_1)$, and $\beta_1$ is the momentum factor.

Similar to our first-order ROSE, we sort the magnitude of the second-order risks $\mathbf{r}_t^{\text{S}}$ in ascending order and calculate the second-order mask $\mathbf{M}_t^{\text{S}}$ with Eq. \ref{eq:3}.
% Unlike first-order ROSE, the gradients with regard to $\mathcal{L}_t^\text{SCE}$ will not be discarded.

\subsection{Ensemble ROSE}
\label{sec:sparse-ensemble}
Since our first-order and second-order ROSE emphasize different kinds of robust parameters, we then propose an ensemble method to aggregate the benefits of the above two mechanisms.

At the $t$-step, we first calculate both the first-order risks $\mathit{r}_t^{\text{F},1},\cdots,\mathit{r}_t^{\text{F},n}$ with Eq. \ref{eq:2} and second-order risks $\mathit{r}_t^{\text{S},1},\cdots,\mathit{r}_t^{\text{S},n}$ with Eq. \ref{eq:4}.
Then we sort both of them in ascending order.
Given upper thresholds $c_h^\text{F}$ and $c_h^\text{S}$, we can compute the first-order and second-order masks as: $\mathbf{M}_t^\text{F}$ and $\mathbf{M}_t^\text{S}$, respectively.
Finally, the ensemble mask $\mathbf{M}_t^\text{E}$ at $t$-step is computed as:
\begin{eqnarray}
\label{eq:5}
    \mathbf{M}_t^\text{E} = \gamma\mathbf{M}_t^\text{F} + (1-\gamma)\mathbf{M}_t^\text{S},
\end{eqnarray}
where $\gamma\in(0, 1)$ is a scaling coefficient hyper-parameter to control the weight of two masks.
\section{Experiment}

\subsection{Datasets} 
We demonstrate the effectiveness of our method using four tasks from GLUE \cite{DBLP:conf/iclr/WangSMHLB19} and AdvGLUE \cite{DBLP:journals/corr/abs-2111-02840} benchmarks.
The General Language Understanding Evaluation (GLUE) is a widely-used benchmark, including $9$ natural language understanding tasks.
The Adversarial GLUE (AdvGLUE) is a robustness benchmark that was created by applying $14$ textual adversarial attack methods to GLUE tasks.
The AdvGLUE adopts careful systematic annotations to curate high-quality and reliable adversarial examples.
We do not employ automatic adversarial attack algorithms to evaluate the adversarial robustness, since they are prone to generating invalid or ambiguous adversarial examples \cite{10.1145/3502223.3502237}.

\textbf{SST-2}~\cite{DBLP:conf/emnlp/SocherPWCMNP13} is a sentiment classification task with single-sentence inputs, which is collected from movie reviews.

\textbf{RTE}~\cite{bentivogli2009fifth} is a natural language inference task constructed based on news and Wikipedia text, aggregated from a series of data from annual textual entailment challenges.

\textbf{QNLI}~\cite{DBLP:conf/emnlp/RajpurkarZLL16} is a task to determine whether the context sentence contains the answer to the question. 

\textbf{QQP}$\footnote{https://quoradata.quora.com/First-Quora-Dataset-Release-Question-Pairs}$~\cite{chen2018quora} is a widely used benchmark involving detecting semantic similarity. 
Each pair is annotated with a binary label indicating whether the two texts are paraphrases or not. 

\subsection{Baselines}
We adopt pre-trained checkpoints of RoBERTa$\rm_{\texttt{BASE}}$ and RoBERTa$\rm_{\texttt{LARGE}}$ \cite{DBLP:journals/corr/abs-1907-11692} as the basis of our experiments.
Besides the vanilla fine-tuning method, we select several suitable baselines for comparison including: 

\textbf{R-Drop} \cite{NEURIPS2021_5a66b920} is a generic regularization strategy, which forces the outputs of two sub-models generated by dropout to be consistent with each other. 
ROSE borrows the idea of dropout twice, but does not fine-tune all parameters to constrain the divergence between two outputs.

\textbf{CHILD-TUNING$\rm_{\textit{D}}$} \cite{DBLP:conf/emnlp/XuLZTCHH21} is a fine-tuning technique, which only updates the most informative subset of parameters of large pre-trained models during the backward process.
Although both CHILD-TUNING and ROSE mask out gradients in the backward process, the specific parameters they update are completely different.

\textbf{SMART} \cite{DBLP:conf/acl/JiangHCLGZ20} is an adversarial training approach, which uses smoothness-inducing regularization and Bregman proximal point optimization during fine-tuning. 

\textbf{FreeLB} \cite{DBLP:conf/iclr/ZhuCGSGL20} is an adversarial training approach built on the top of language models, which improves higher invariance in word embedding space and minimizes the resultant adversarial loss around input samples.

% The last two models are state-of-the-art adversarial training methods.
% We name our method as "ROSE-First", "ROSE-Second" and "ROSE-Ensemble" corresponding to the three variants, respectively.

\subsection{Experimental Settings}
\label{sec:4.3}
Our implementation of ROSE is based on Huggingface library$\footnote{ https://github.com/huggingface/transformers}$~\cite{wolf-etal-2020-transformers}. 
Batch size for RTE is set to $16$, and for other tasks it is set to $32$.
Dropout rates are all set to $0.1$.
We carry out grid search of learning rate $\in \{1e-5, 2e-5, \cdots, 1e-4 \}$ and upper threshold $\in \{10\%,20\%, \cdots, 90\% \}$.
For ROSE-ensemble, we simply set $\gamma=0.5$ in Eq. \ref{eq:5}.
The maximum number of epochs is set to $10$.
For the replication study, we report the average accuracy over $5$ random seeds on the GLUE and AdvGLUE development sets after fine-tuning the pre-trained models on the corresponding GLUE training data.

For all the baselines, we either perform grid search or adopt the parameter combination provided by the official codebase to select the best parameters.
Similarly, we report the average results on two benchmarks over $5$ random seeds using the same evaluation schema.
% More details are provided in Appendix \ref{sec:appendix}.

\subsection{Main Results}
\label{sec:main-result}

\begin{table*}[thp]
\centering
\resizebox{\linewidth}{!}{
\begin{tabular}{l|cc|cc|cc|cc|cc}
\toprule
\multirow{2}{*}{\textbf{Model}} & \multicolumn{2}{c|}{\textbf{SST-2}} & \multicolumn{2}{c|}{\textbf{RTE}} & \multicolumn{2}{c|}{\textbf{QNLI}} & \multicolumn{2}{c|}{\textbf{QQP}} & \textbf{Avg} & \textbf{Avg} \\
& \small GLUE & \small AdvGLUE & \small GLUE & \small AdvGLUE & \small GLUE & \small AdvGLUE & \small GLUE & \small AdvGLUE & \small GLUE$+$AdvGLUE & \small $\Delta\downarrow$\\
\hline
\hline
\multicolumn{11}{c}{\textbf{RoBERTa}$\rm_{\texttt{BASE}}$} \\
\hline
Vanilla & 94.29 & 24.05 & 77.91 & 28.15 & 92.97 & 27.43 & 91.58 &	19.49 & 56.98 & 64.41  \\
R-Drop & 95.32 & 27.84 & 79.86 & 31.36 & 93.30 & 28.92 & 91.86 & 37.44 & 60.74 & 58.70   \\
CHILD-TUNING$\rm_\textit{D}$ & 94.21 & 23.82 & 75.52 & 16.54 & 92.36 & 31.89 & 91.64 & 17.95 & 55.49 & 65.88 \\
SMART & 94.98 & 35.95 & 77.54 & 24.44 & 93.35 & 34.29 & 91.04 & 46.58 & 62.27 & 53.91   \\
FreeLB & 94.89 & 35.81 & 78.42 & 32.10 & 93.12 & 36.22 & 92.04 & 44.10 & 63.34 & 52.56 \\
\hline
ROSE-First & 94.84 & 37.67 & 78.34 & 35.49 & 92.19 & 44.19 & 89.56 & 44.44 & 64.59 & 48.29 \\
ROSE-Second & 93.78 & 36.99 & 78.16 & 37.97 & 92.41 & 34.63 & 90.48 & 45.73 & 63.77 & 49.88   \\
ROSE-Ensemble & 94.09 & 39.36 & 78.63 & 38.02 & 92.64 & 39.59 & 90.39 & 47.44 & \textbf{65.02} & \textbf{47.84}   \\
\hline
\hline
\multicolumn{11}{c}{\textbf{RoBERTa}$\rm_{\texttt{LARGE}}$} \\
\hline
Vanilla & 96.08 & 56.08 & 85.92 & 61.73 & 94.58 & 63.38 & 92.09 & 40.60 & 73.81 & 36.72   \\
R-Drop & 96.59 & 53.38 & 85.56 & 66.67 & 95.01 & 55.95 & 92.35 & 44.80 & 73.79 & 37.18   \\
CHILD-TUNING$\rm_\textit{D}$ & 95.91 & 51.35 & 85.92 & 61.73 & 94.30 & 58.11 & 92.03 & 43.59 & 72.87 & 38.35   \\
SMART & 96.67 & 59.12 & 85.02 & 69.14 & 94.91 & 61.04 & 92.12 & 50.85 & 76.11 & 32.14   \\
FreeLB & 96.49 & 59.32 & 86.76 & 66.91 & 94.99 & 62.30 & 92.60 & 48.21 &	75.95 & 33.53   \\
\hline
ROSE-First & 95.58 &	57.77 &	85.13 &	70.62 &	94.08 &	64.02 &	90.67 &	60.26 &	77.27 & 28.20 \\
ROSE-Second & 96.29 &	60.59 &	85.08 &	67.49 &	94.72 &	63.68 &	91.68 &	55.90 &	76.93 & 30.03  \\
ROSE-Ensemble & 96.10 &	60.81 &	85.92 &	71.11 &	94.26 &	64.64 &	91.46 &	60.51 &	\textbf{78.10} & \textbf{27.67}  \\
\bottomrule
\end{tabular}}
\caption{Model performance on GLUE and AdvGLUE benchmarks. The results are accuracy averaged over 5 random seeds. 
All values are reported by percentage (\%). 
We also report the macro-average of per-task scores. 
The last column is the drop from GLUE to AdvGLUE, the smaller the better.
The \textbf{bold} represents ROSE is significantly better (1-tailed t-test, p-value $< 0.05$) than the baselines.}
\label{tab:main-result}
\end{table*}

We compare ROSE-First, ROSE-Second, and ROSE-Ensemble to all baselines on SST-2, RTE, QNLI, and QQP tasks from GLUE and AdvGLUE benchmarks. 
The overall results are summarized in Table \ref{tab:main-result}.
We observe that:

(1) Our proposed ROSE substantially improves the robustness of fine-tuned pre-trained language models, while maintaining competitive benign performances to previous methods.
Despite the effectiveness of ROSE-First and ROSE-Second varies on tasks and model size, both of them surpass the existing methods.
ROSE-Ensemble aggregates the advantages of first-order and second-order strategies, providing the strongest adversarial robustness.
In particular, ROSE-Ensemble$\rm_{\texttt{BASE}}$ outperforms vanilla RoBERTa$\rm_{\texttt{BASE}}$ model by $8.04\%$ average score.
ROSE-Ensemble$\rm_{\texttt{LARGE}}$ beats RoBERTa$\rm_{\texttt{LARGE}}$ by $4.29\%$ on average.

(2) ROSE consistently outperforms CHILD-TUNING$\rm_{\textit{D}}$ and R-Drop, which both share some similarities with our method.
CHILD-TUNING$\rm_{\textit{D}}$, which masks out the most inessential parameters in the backward process, shows the worst robustness on most datasets.
R-Drop uses dropout to regularize the output of models.
Results indicate that R-Drop improves robustness on a number of tasks, but it is not competitive with strong defense methods.
We will explore the effectiveness of our robust selection strategies further in Section \ref{sec:3.5}.

(3) Our method also surpasses the two strong baselines SMART and FreeLB, which employ the most prevalent adversarial training idea to improve the robustness of pre-trained models.
For instance, ROSE-Ensemble$\rm_{\texttt{BASE}}$ yields an improvement of up to $2.75\%$ average score over SMART$\rm_{\texttt{BASE}}$.
ROSE-Ensemble gains $2.15\%$ average score improvement compared to FreeLB with RoBERTa$\rm_{\texttt{LARGE}}$.
Furthermore, SMART and FreeLB are both inefficient and heavily correlated with model structure, while our ROSE does not suffer from these issues.

\subsection{Extensions of ROSE to Existing Method}
\label{sec:3.5}

\begin{table*}[htbp]
\centering
\resizebox{\linewidth}{!}{
\begin{tabular}{l|cc|cc|cc|cc|cc}
\toprule
\multirow{2}{*}{\textbf{Model}} & \multicolumn{2}{c|}{\textbf{SST-2}} & \multicolumn{2}{c|}{\textbf{RTE}} & \multicolumn{2}{c|}{\textbf{QNLI}} & \multicolumn{2}{c|}{\textbf{QQP}} & \textbf{Avg} & \textbf{Avg} \\
& \small GLUE & \small AdvGLUE & \small GLUE & \small AdvGLUE & \small GLUE & \small AdvGLUE & \small GLUE & \small AdvGLUE & \small GLUE$+$AdvGLUE & \small $\Delta\downarrow$\\
\hline
\hline
\multicolumn{11}{c}{\textbf{RoBERTa}$\rm_{\texttt{BASE}}$} \\
\hline
R-Drop & 95.32 & 27.84 & 79.86 & 31.36 & 93.30 & 28.92 & 91.86 & 37.44 & 60.74 & 58.70   \\
\hline
$+$ ROSE-First & 94.77 & 32.30 & 78.34 & 36.05 & 92.92 & 39.46 & 90.62 & 42.82 & 63.41 & 51.51  \\
$+$ ROSE-Second & 95.05 & 30.00 & 73.93 & 36.79 & 92.60 & 32.03 & 91.96 & 39.23 & 61.45 & 53.87  \\
$+$ ROSE-Ensemble & 92.36 & 35.68 & 77.83 & 36.54 & 92.50 & 40.27 & 90.90 & 42.56 & \textbf{63.58} & \textbf{49.64}  \\
\hline
\hline
\multicolumn{11}{c}{\textbf{RoBERTa}$\rm_{\texttt{LARGE}}$} \\
\hline
R-Drop & 96.59 & 53.38 & 85.56 & 66.67 & 95.01 & 55.95 & 92.35 & 44.80 & 73.79 & 37.18    \\
\hline
$+$ ROSE-First & 96.54 & 55.68 & 85.85 & 68.89 & 94.38 & 56.62 & 91.95 & 53.21 & 75.39 & 33.58   \\
$+$ ROSE-Second & 96.63 & 55.81 & 85.27 & 66.91 & 94.84 & 57.16  & 92.38 & 50.26 & 74.91 & 34.74.  \\
$+$ ROSE-Ensemble & 96.72 & 56.62 & 85.20 & 69.13 & 94.81 & 57.97 & 92.27 & 54.61  & \textbf{75.92 } & \textbf{32.66}  \\
\bottomrule
\end{tabular}}
\caption{Experiment results for R-Drop incorporated with ROSE. ROSE improves the robustness of R-Drop further.
The \textbf{bold} represents ROSE is significantly better (1-tailed t-test, p-value $< 0.05$) than R-Drop.}
\label{tab:ablation}
\end{table*}

ROSE is a generic method and can be easily incorporated into other well-recognized methods.
In this section, we incorporate ROSE into R-Drop and examine whether it is still effective.
Since the optimization objective of R-Drop is a weighted sum of softmax cross-entropy loss and KL-divergence, we decouple them from the aggregated loss and use them to perform our first-order and second-order mask calculations, respectively.
Noted that, in the backward process we still use gradient calculated with regards to the aggregated loss to update, which is different from the ROSE process.

We primarily adopt the best parameter combinations from the main experiments in Section \ref{sec:main-result}, including the learning rates and upper thresholds.
We follow the settings from R-Drop for other parameters.
We conduct experiments using both RoBERTa$\rm_{\texttt{BASE}}$ and RoBERTa$\rm_{\texttt{LARGE}}$.

Results are shown in Table \ref{tab:ablation}.
Generally, ROSE improves the adversarial robustness of R-Drop by a large margin, and maintains competitive benign performances at the same time.
For example, ROSE-First$\rm_{\texttt{BASE}}$ promotes the adversarial robustness of R-Drop$\rm_{\texttt{BASE}}$ on QNLI task from $28.92\%$ to $39.46\%$.
R-Drop patched with ROSE-Second witnesses an improvement on QQP task from $44.80\%$ to $50.26\%$ using RoBERTa$\rm_{\texttt{LARGE}}$.
Notably, our ROSE-Ensemble outperforms R-Drop by roughly $3$ points on average for both model sizes.
The above results indicate that when incorporated into existing methods, ROSE can enhance their adversarial robustness even further.

\subsection{Effect of Scalar $\gamma$}
\label{sec:3.6}
Further, we investigate the impact of the scaling coefficient $\gamma$ in our ROSE-Ensemble.
Here we vary the $\gamma\in\{0.1, 0.3, 0.5, 0.7, 0.9\}$ and conduct experiments on four tasks, where $\gamma=0.5$ is the current setting.
We adopt the setting from Section \ref{sec:main-result} for other parameters. 

\begin{table}[htbp]
    \centering
    \resizebox{\linewidth}{!}{
    \begin{tabular}{l|c|c|c|c}
    \toprule
    & \multicolumn{2}{c|}{\textbf{RoBERTa}$\rm_{\texttt{BASE}}$} & \multicolumn{2}{c}{\textbf{RoBERTa}$\rm_{\texttt{LARGE}}$} \\
    \cline{2-5}
    \multirow{3}{*}[2ex]{\textbf{$\gamma$}} & \textbf{Avg} & \textbf{Avg} & \textbf{Avg} & \textbf{Avg} \\
    & \small GLUE$+$AdvGLUE & \small $\Delta\downarrow$ & \small GLUE$+$AdvGLUE & \small $\Delta\downarrow$ \\
    \hline
    $0.1$ & 59.35 & 56.81 & 75.07 & 28.99 \\
    $0.3$ & 58.42 & 55.13 & 75.51 & 30.27 \\
    $0.5$ & 65.02 & 47.84 & 78.10 & 27.67 \\
    $0.7$ & 61.09 & 51.36 & 75.60 & 30.52 \\
    $0.9$ & 59.88 & 50.82 & 73.22 & 32.25 \\
    \bottomrule
    \end{tabular}}
    \caption{Results for ROSE-Ensemble with different $\gamma$.}
    \label{tab:gamma-effect}
\end{table}

The results are presented in Table \ref{tab:gamma-effect}.
It is shown that the best-balanced choice is $\gamma=0.5$, but ROSE-Ensemble can stably improve the robustness using other $\gamma$.
Furthermore, ROSE achieves more substantial performance when applied to pre-trained language models of greater complexity.
% Therefore the value of $\gamma$ should be chosen carefully according to the nature of specific tasks.
\section{Analysis}

In this section, we conduct further analysis to reveal the inner working of ROSE.

\subsection{Two-dimensional Loss Visualization}

The loss landscape \cite{DBLP:journals/corr/GoodfellowV14,DBLP:conf/nips/Li0TSG18} is a valid and effective indicator to characterize the property of neural networks.
It has been empirically demonstrated that flatter and wider optima correlate well with better robustness.
We plot and compare two-dimensional loss landscapes of the solutions found by vanilla fine-tuning and our ROSE.
Visualizations on various tasks show that our ROSE generally leads to flatter and wider optima, thus improving the adversarial robustness of adapted models.

\begin{figure*}[t]
    \centering
    \subfigure{
    \rotatebox{90}{~~~~~~~~~~~~~SST-2}
		\begin{minipage}[t]{0.22\textwidth}
			\includegraphics[width=1\textwidth]{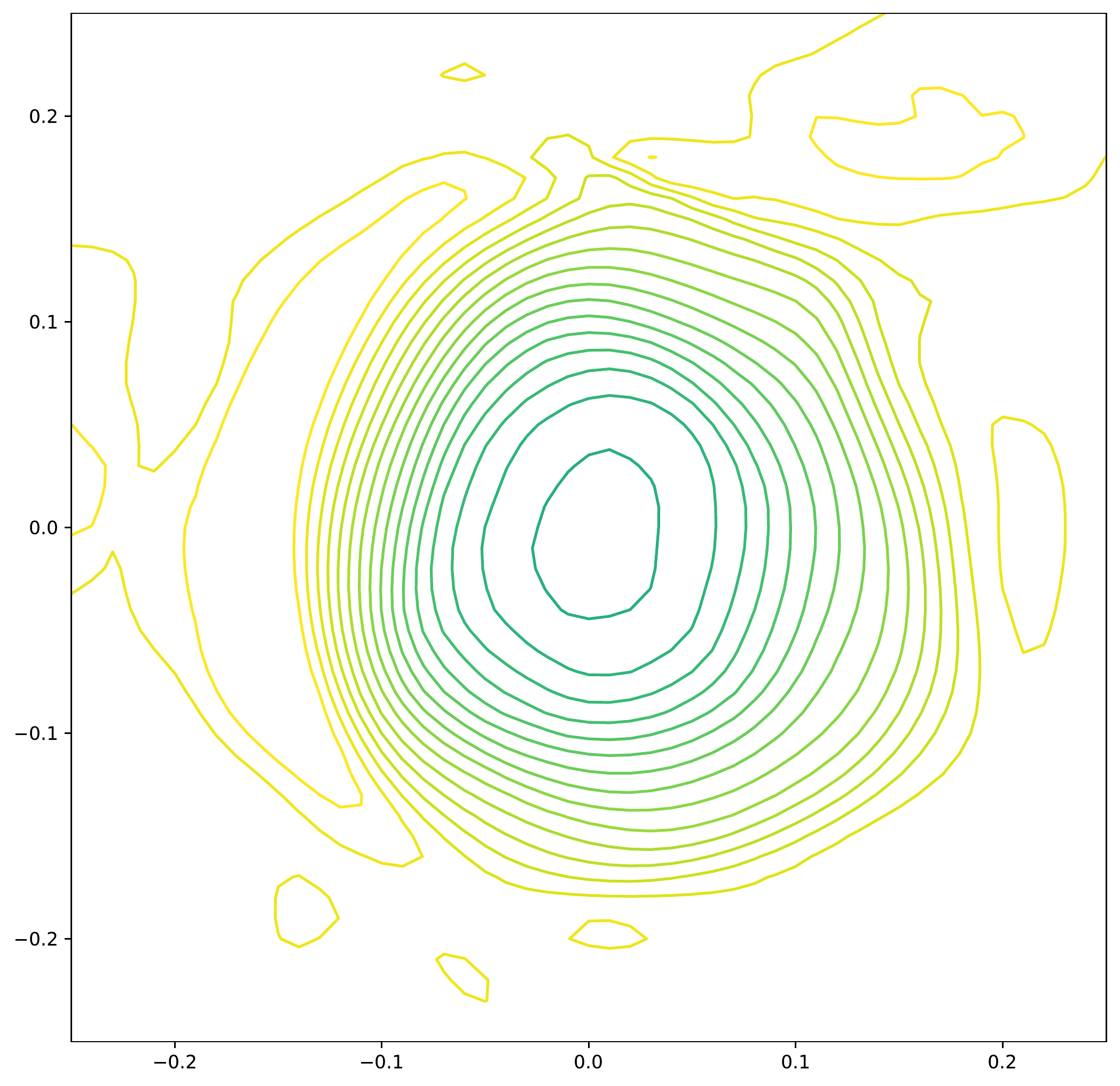}
		\end{minipage}
% 		\label{fig:ll-2D-vanilla}
	}
    \subfigure{
    	\begin{minipage}[t]{0.22\textwidth}
   		\includegraphics[width=1\textwidth]{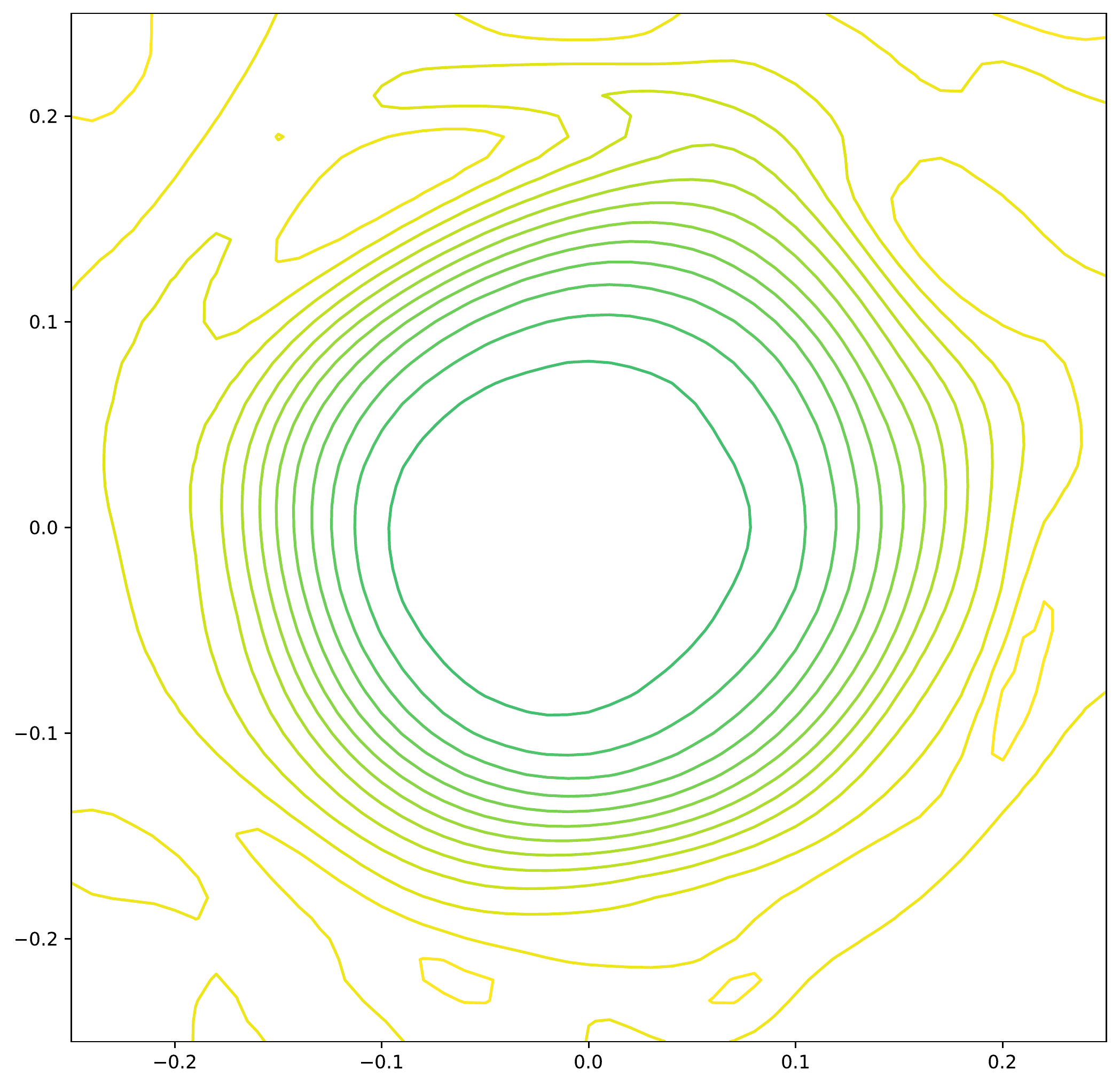}
    	\end{minipage}
% 	\label{fig:ll-2D-first}
    }
    \subfigure{
    	\begin{minipage}[t]{0.22\textwidth}
   		\includegraphics[width=1\textwidth]{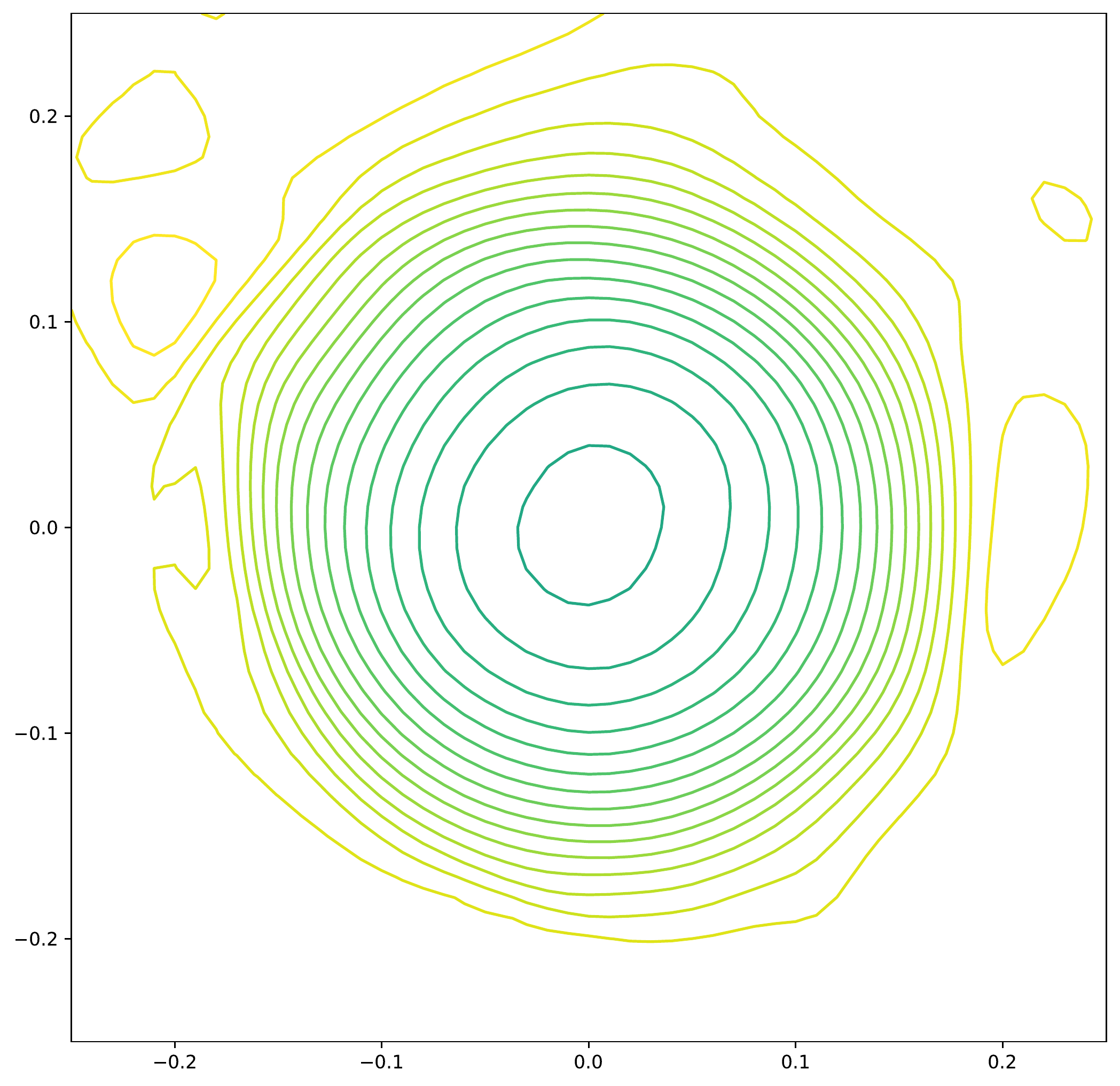}
    	\end{minipage}
% 	\label{fig:ll-2D-second}
    }
    \subfigure{
    	\begin{minipage}[t]{0.22\textwidth}
   		\includegraphics[width=1\textwidth]{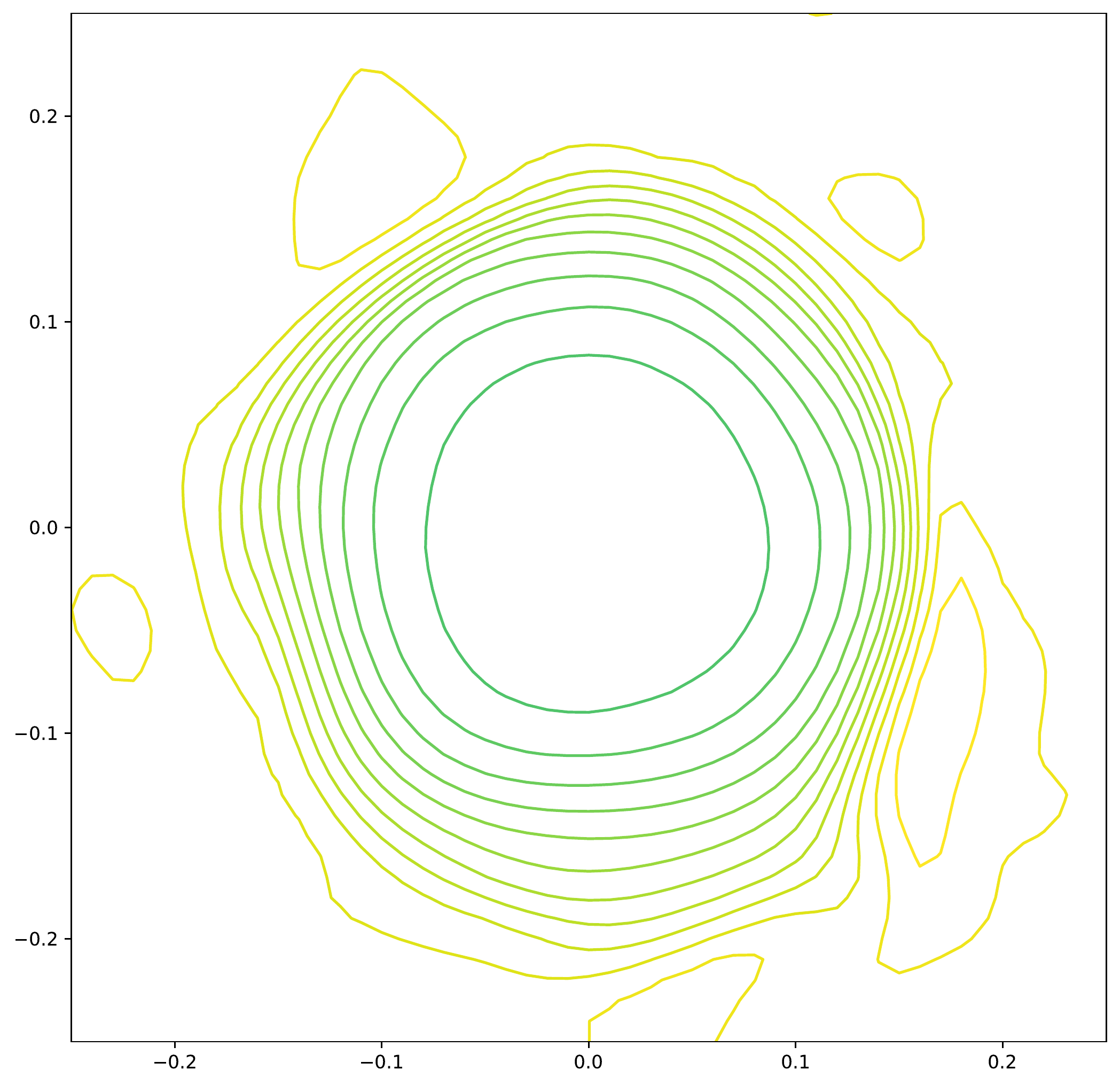}
    	\end{minipage}
% 	\label{fig:ll-2D-ensemble}
    }
    \\
    \vspace{-3mm}
    \subfigure{
    \rotatebox{90}{~~~~~~~~~~~~~RTE}
		\begin{minipage}[t]{0.22\textwidth}
			\includegraphics[width=1\textwidth]{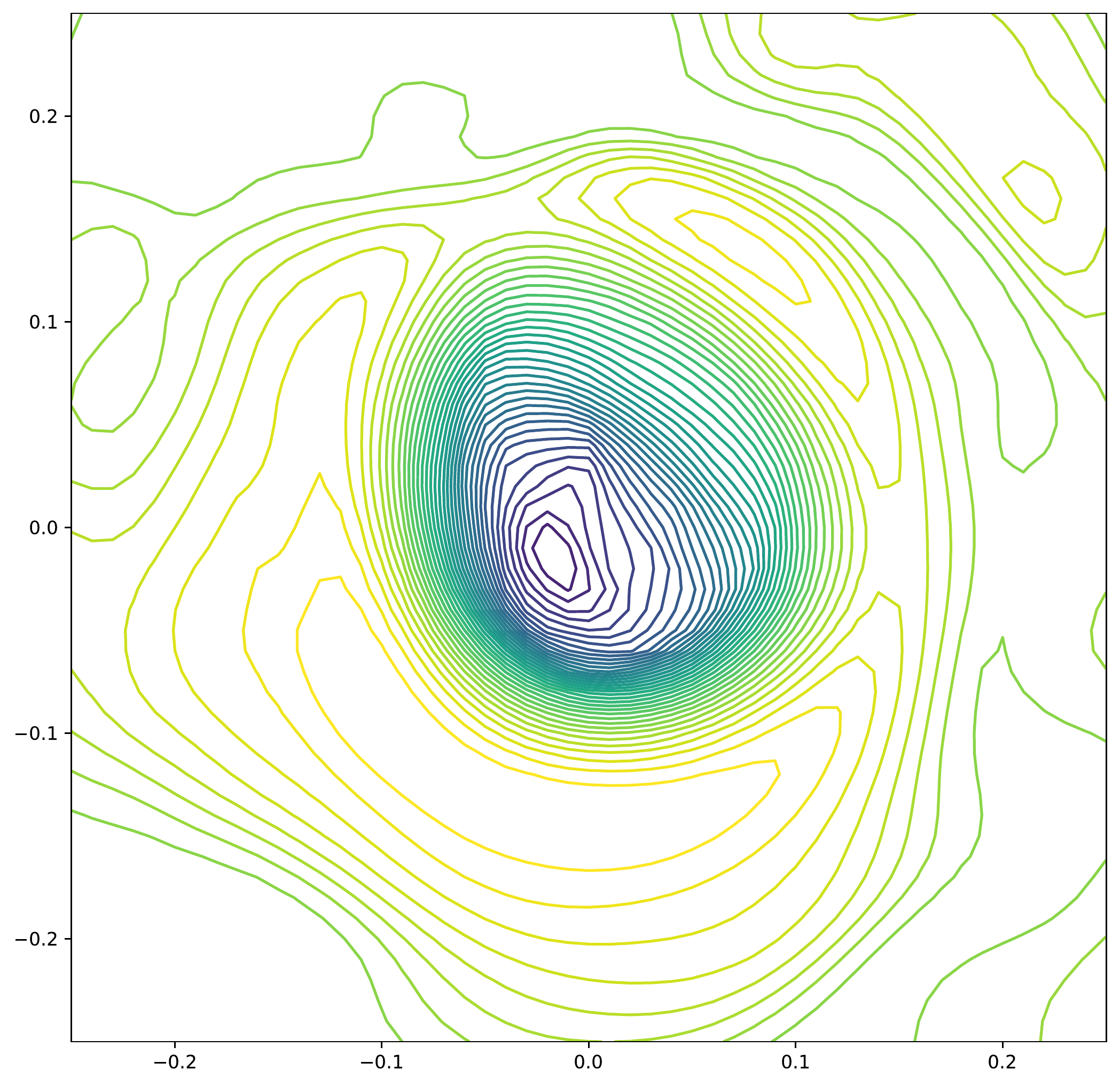}
		\end{minipage}
% 		\label{fig:ll-2D-vanilla}
	}
    \subfigure{
    	\begin{minipage}[t]{0.22\textwidth}
   		\includegraphics[width=1\textwidth]{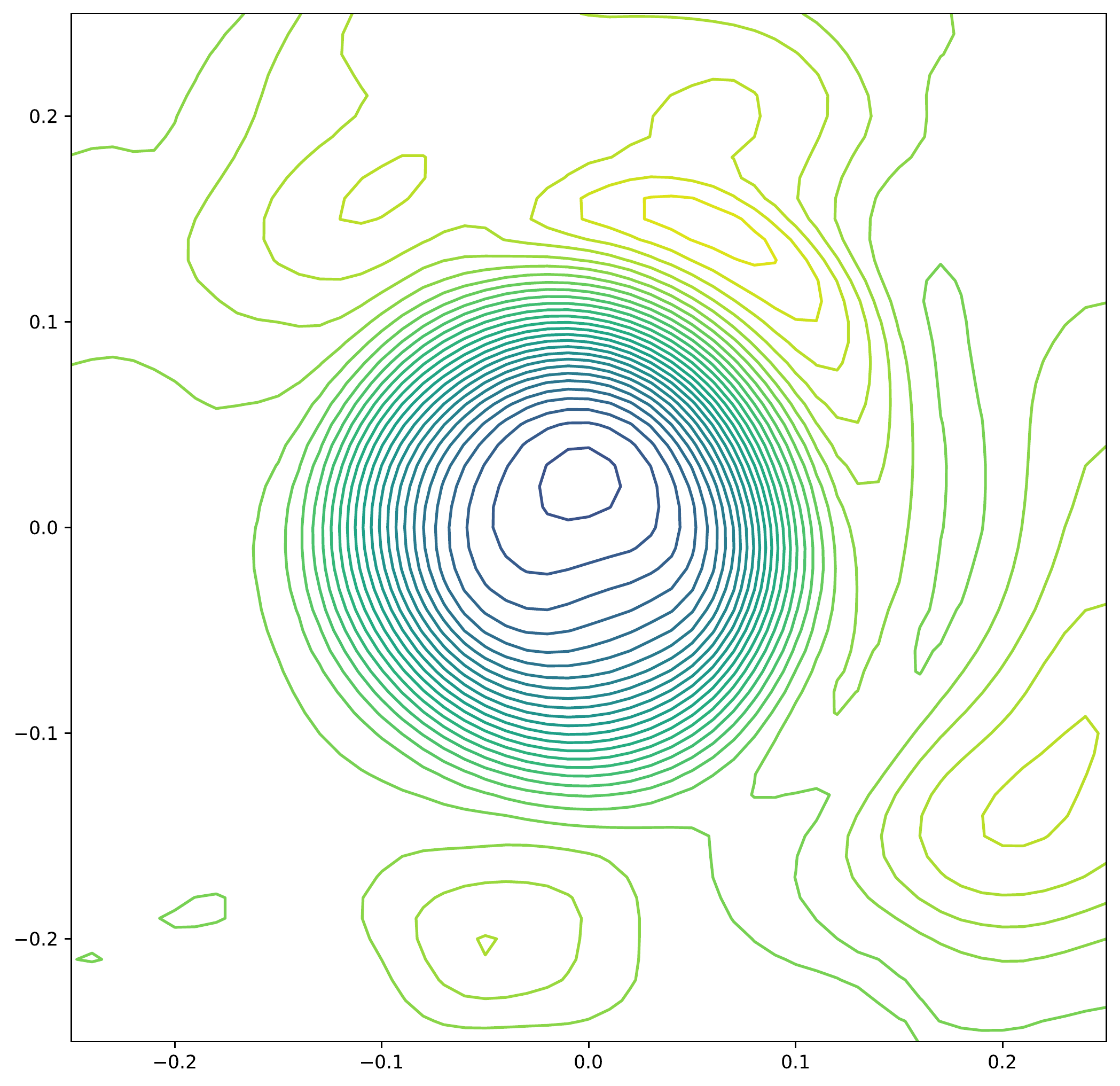}
    	\end{minipage}
% 	\label{fig:ll-2D-first}
    }
    \subfigure{
    	\begin{minipage}[t]{0.22\textwidth}
   		\includegraphics[width=1\textwidth]{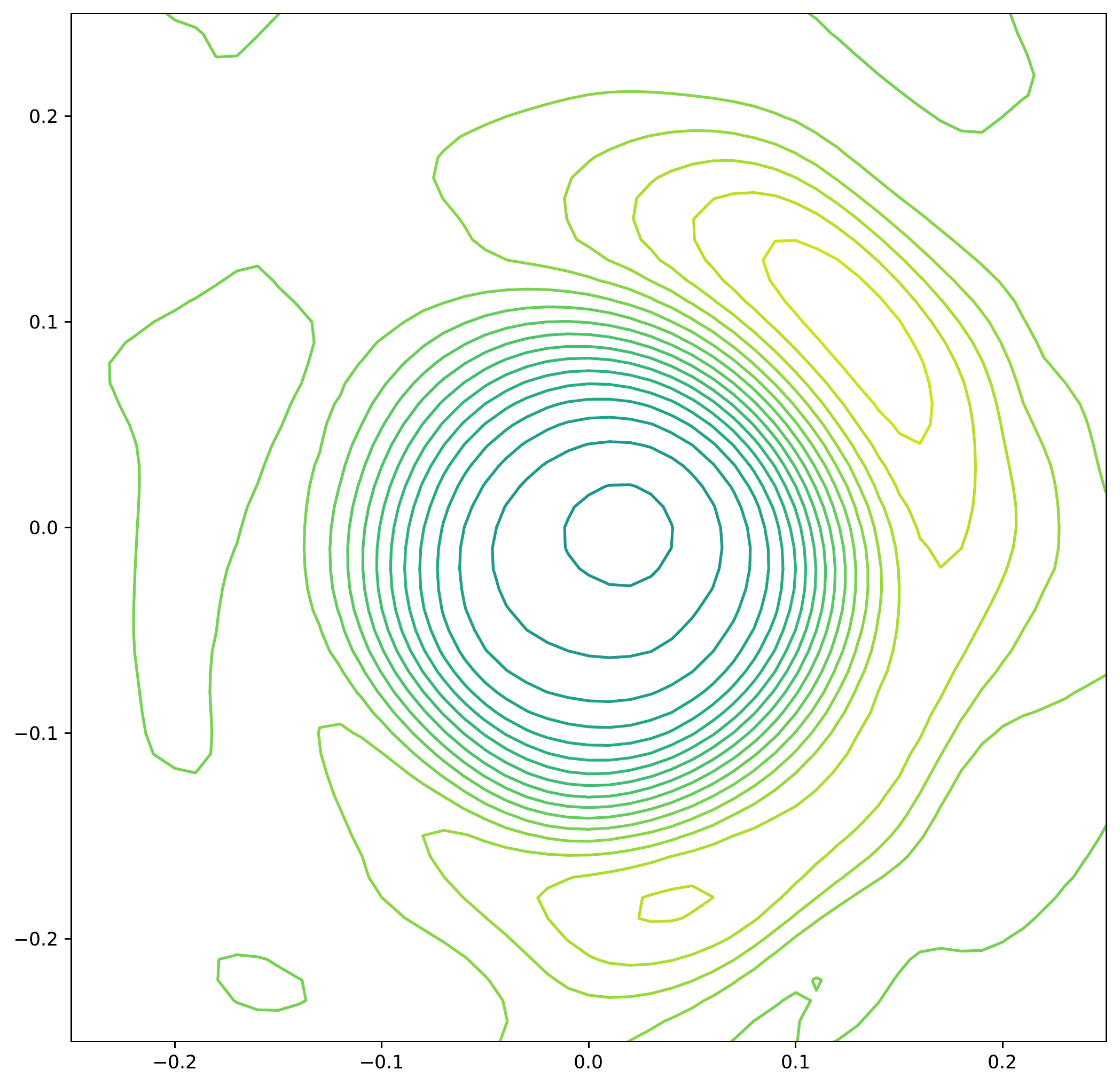}
    	\end{minipage}
% 	\label{fig:ll-2D-second}
    }
    \subfigure{
    	\begin{minipage}[t]{0.22\textwidth}
   		\includegraphics[width=1\textwidth]{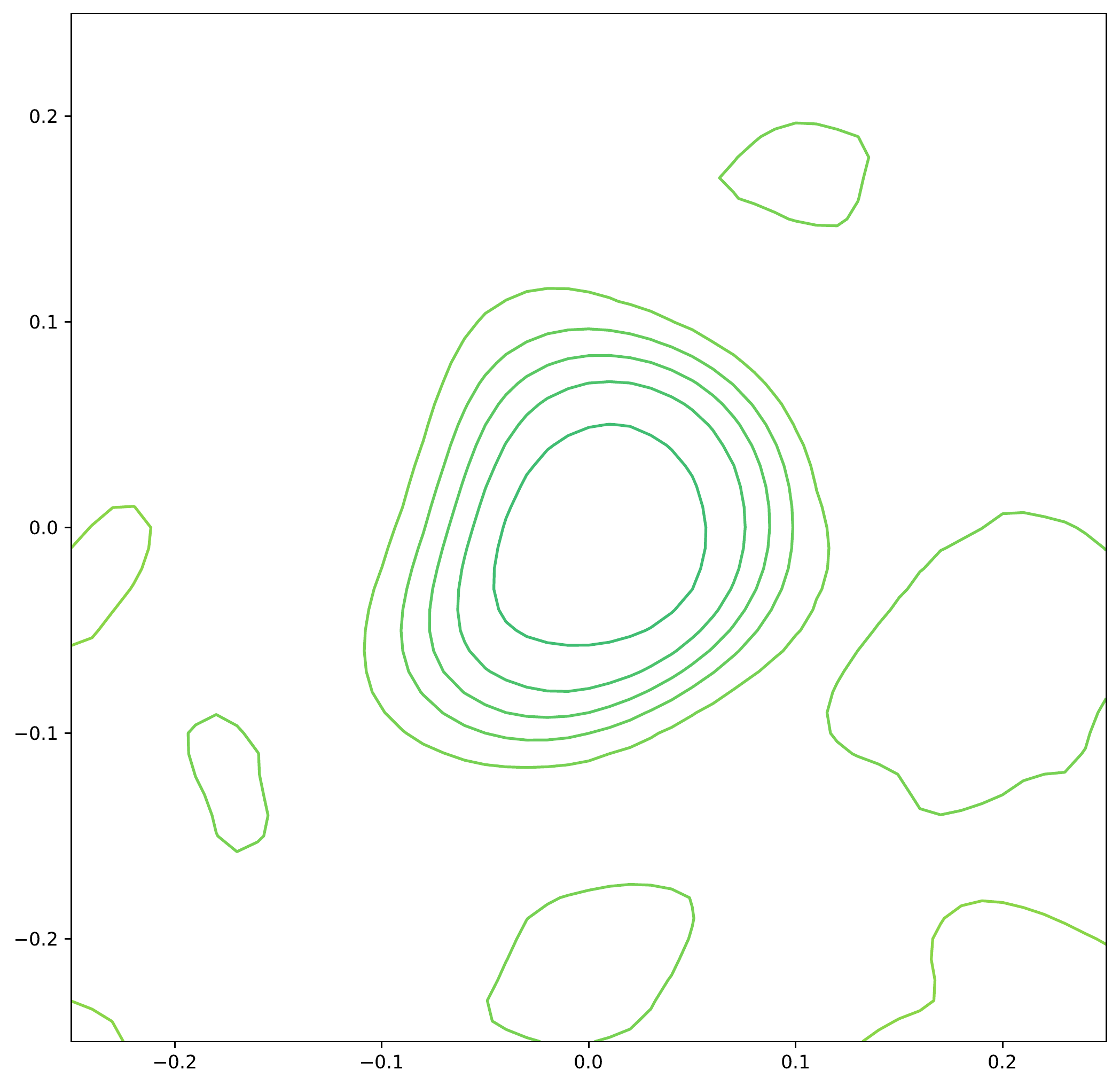}
    	\end{minipage}
% 	\label{fig:ll-2D-ensemble}
    }
    \\
    \vspace{-3mm}
    \subfigure{
    \rotatebox{90}{~~~~~~~~~~~~~QNLI}
		\begin{minipage}[t]{0.22\textwidth}
			\includegraphics[width=1\textwidth]{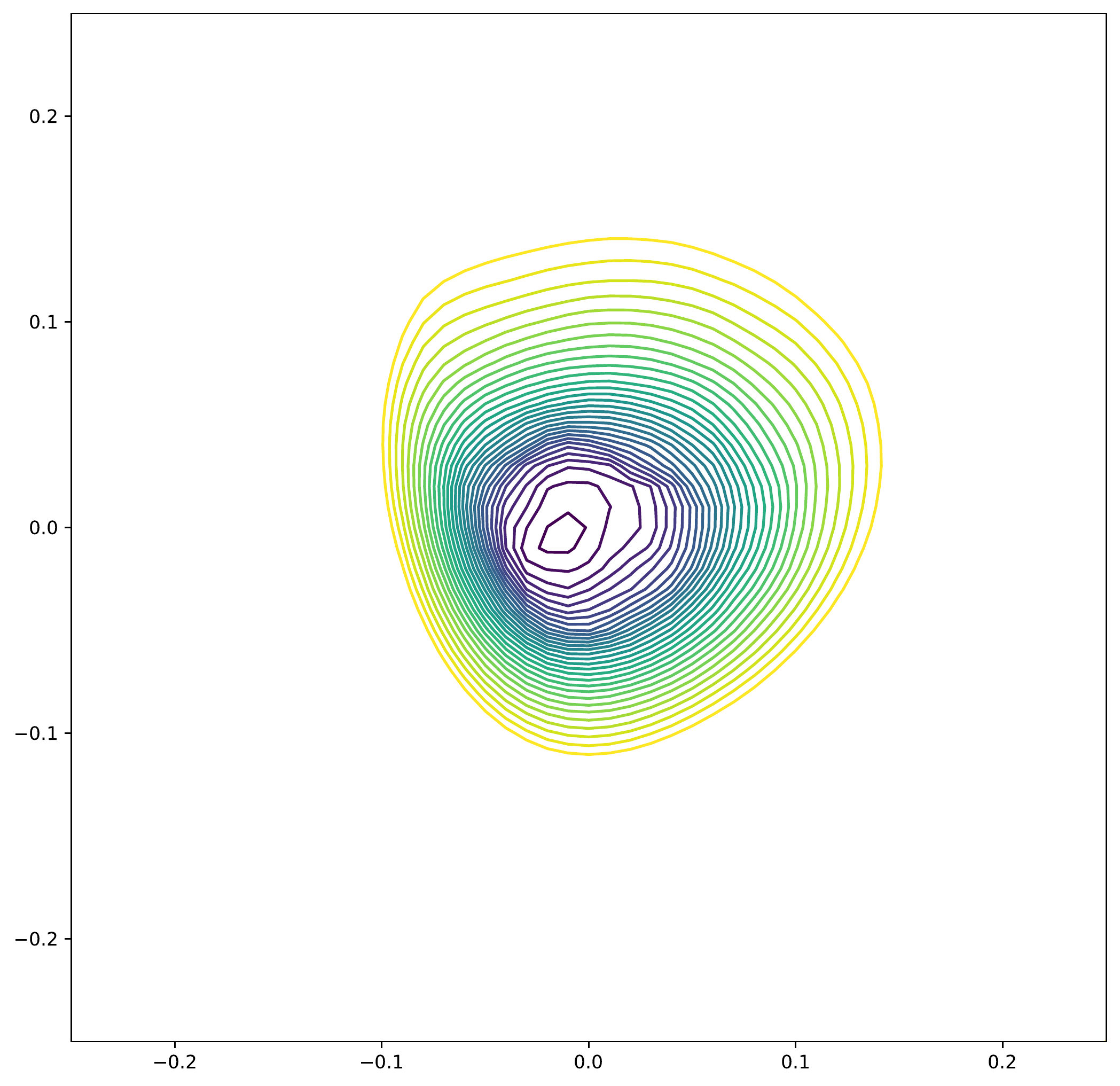}
		\end{minipage}
% 		\label{fig:ll-2D-vanilla}
	}
    \subfigure{
    	\begin{minipage}[t]{0.22\textwidth}
   		\includegraphics[width=1\textwidth]{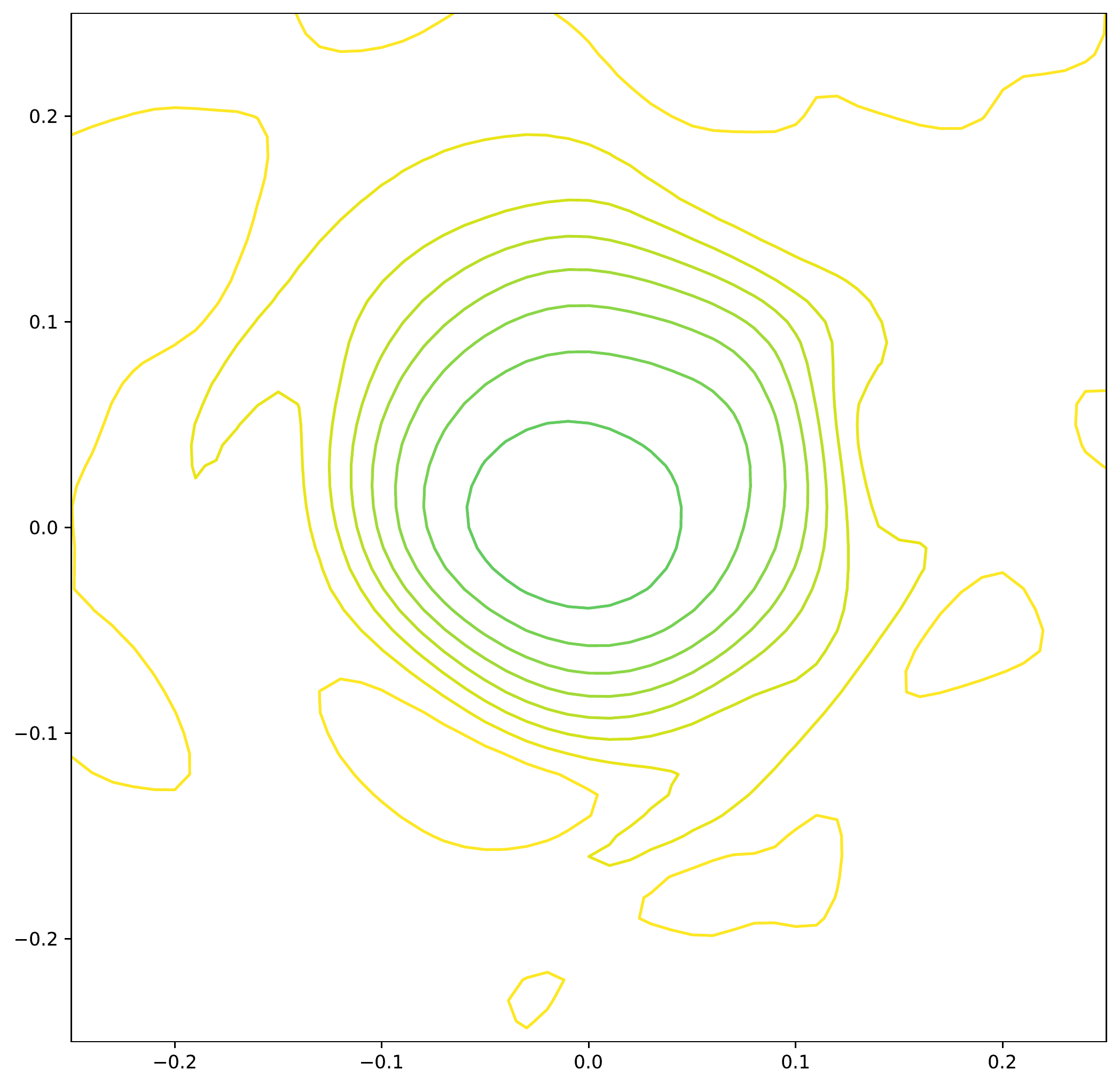}
    	\end{minipage}
% 	\label{fig:ll-2D-first}
    }
    \subfigure{
    	\begin{minipage}[t]{0.22\textwidth}
   		\includegraphics[width=1\textwidth]{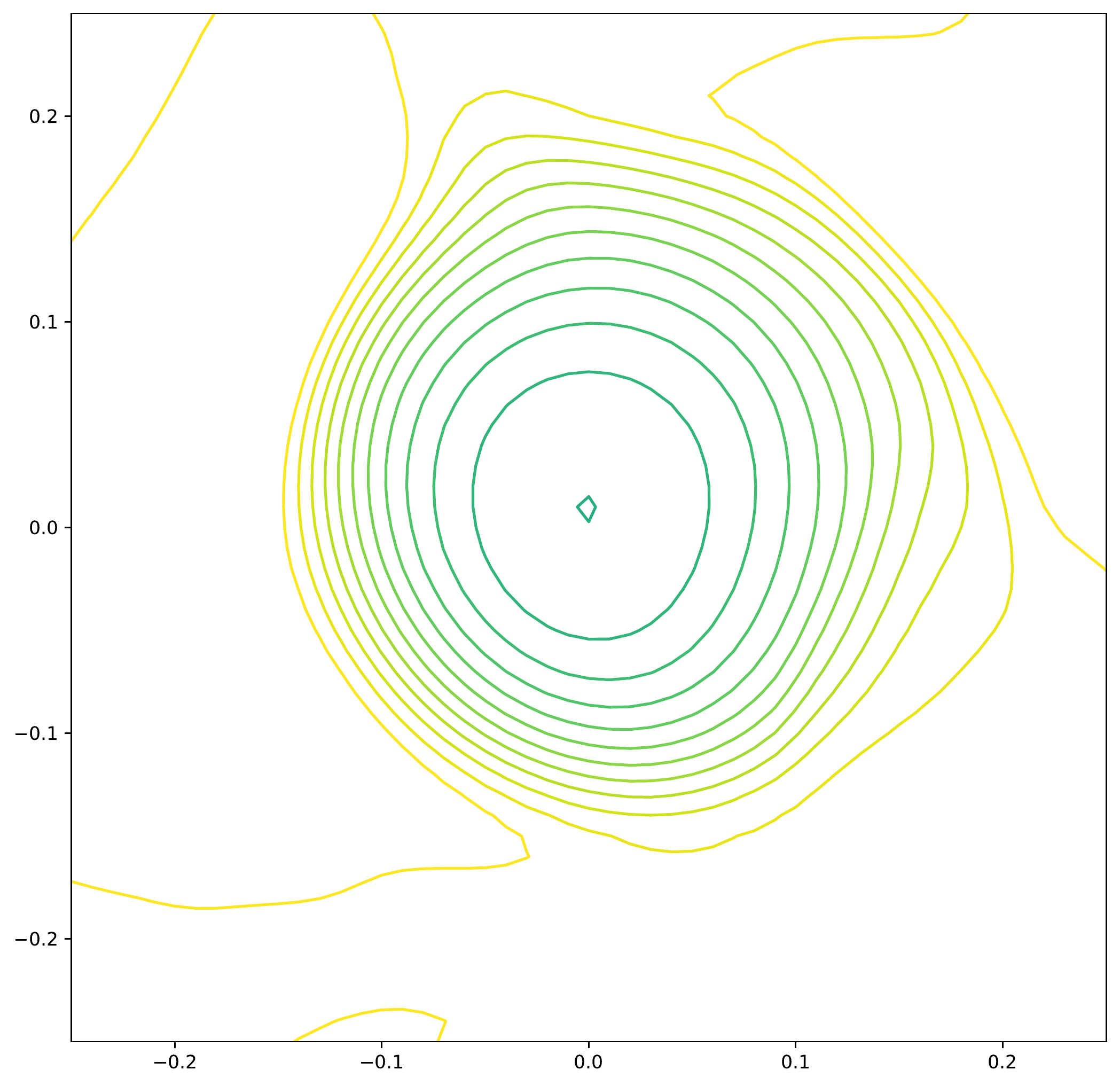}
    	\end{minipage}
% 	\label{fig:ll-2D-second}
    }
    \subfigure{
    	\begin{minipage}[t]{0.22\textwidth}
   		\includegraphics[width=1\textwidth]{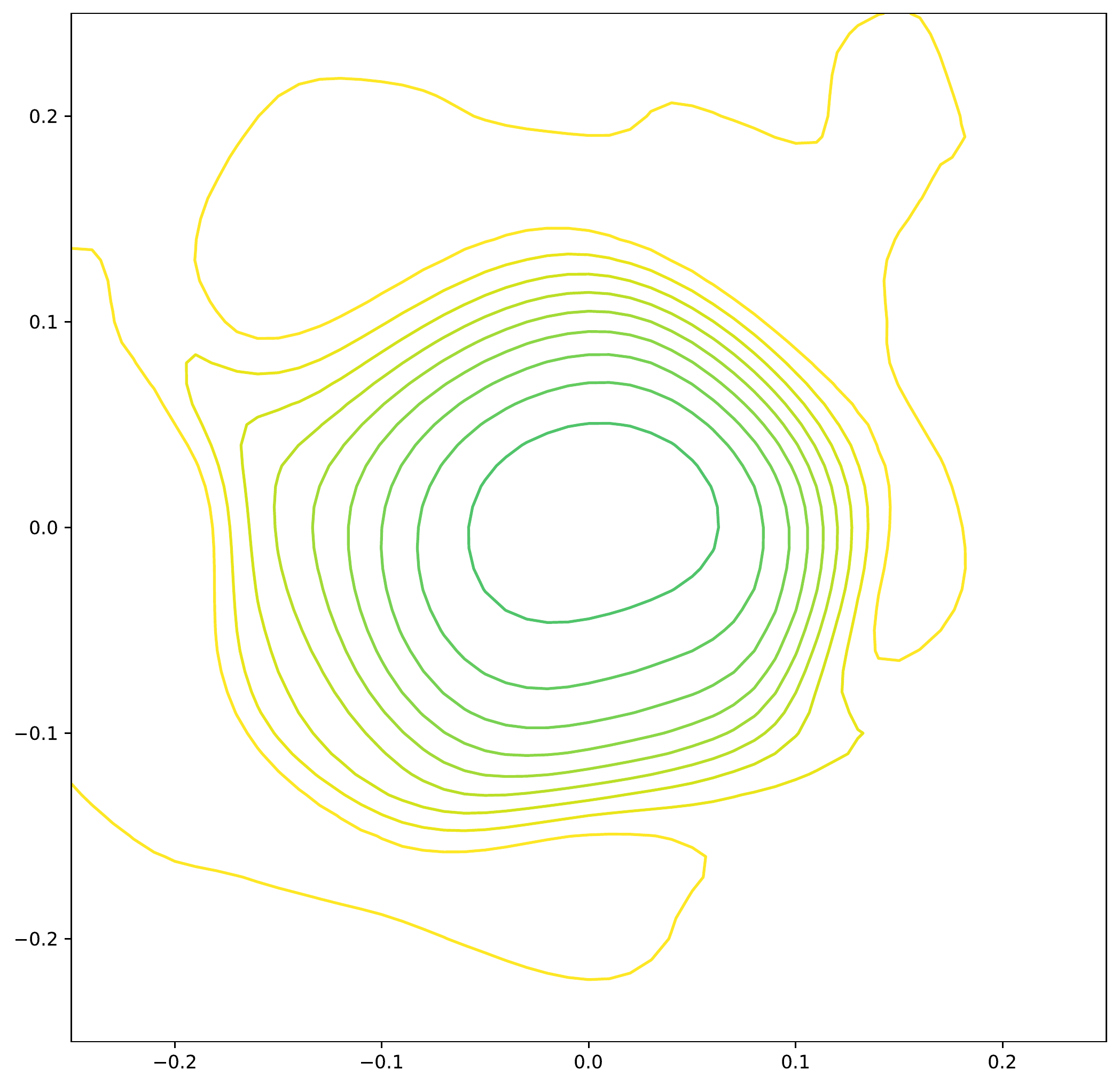}
    	\end{minipage}
% 	\label{fig:ll-2D-ensemble}
    }
    \\
    \vspace{-3mm}
    \setcounter{subfigure}{0}
    \subfigure[Vanilla]{
    \rotatebox{90}{~~~~~~~~~~~~~QQP}
		\begin{minipage}[b]{0.22\textwidth}
			\includegraphics[width=1\textwidth]{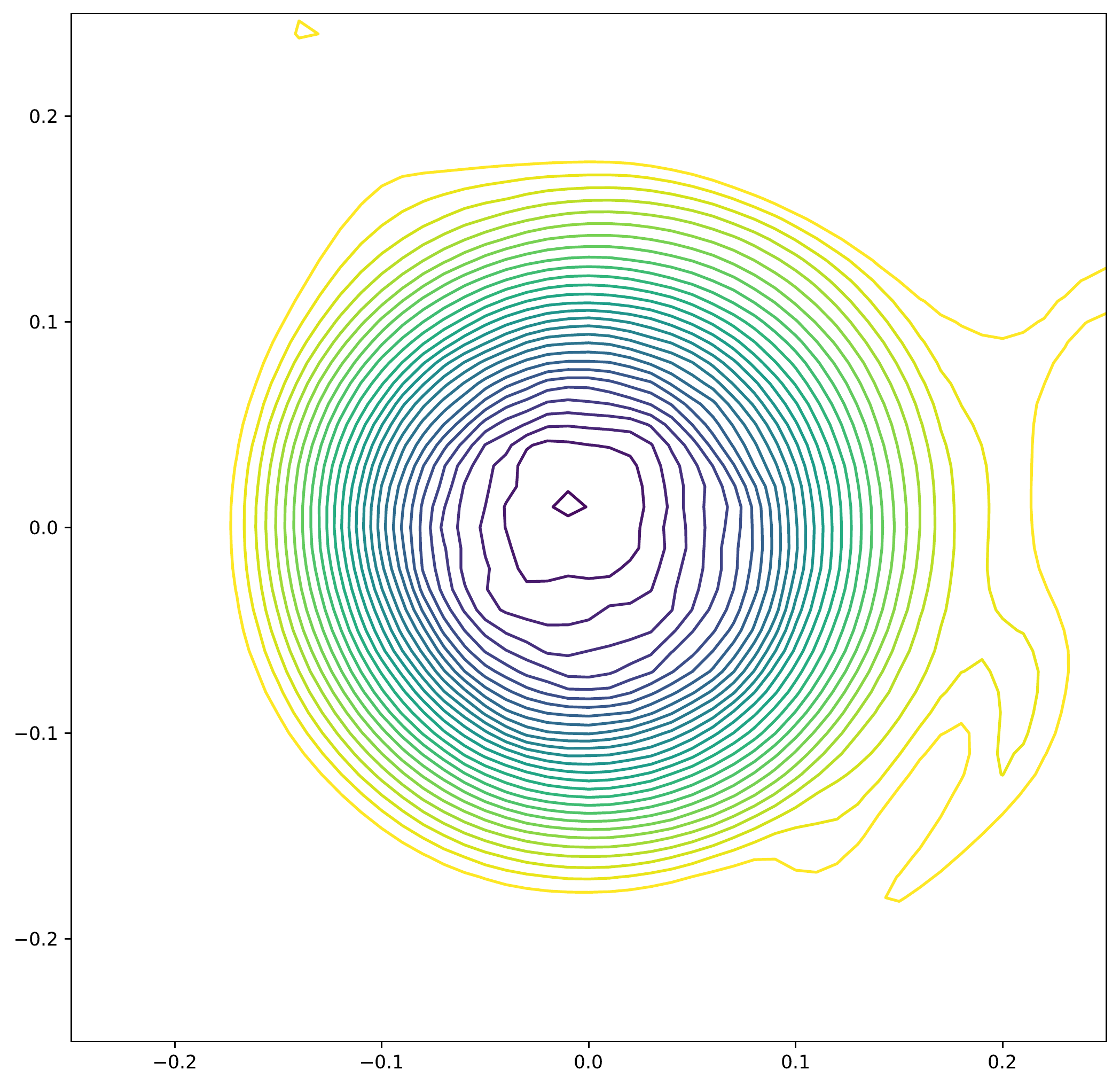}
		\end{minipage}
% 		\label{fig:ll-2D-vanilla}
	}
    \subfigure[ROSE-First]{
    	\begin{minipage}[b]{0.22\textwidth}
   		\includegraphics[width=1\textwidth]{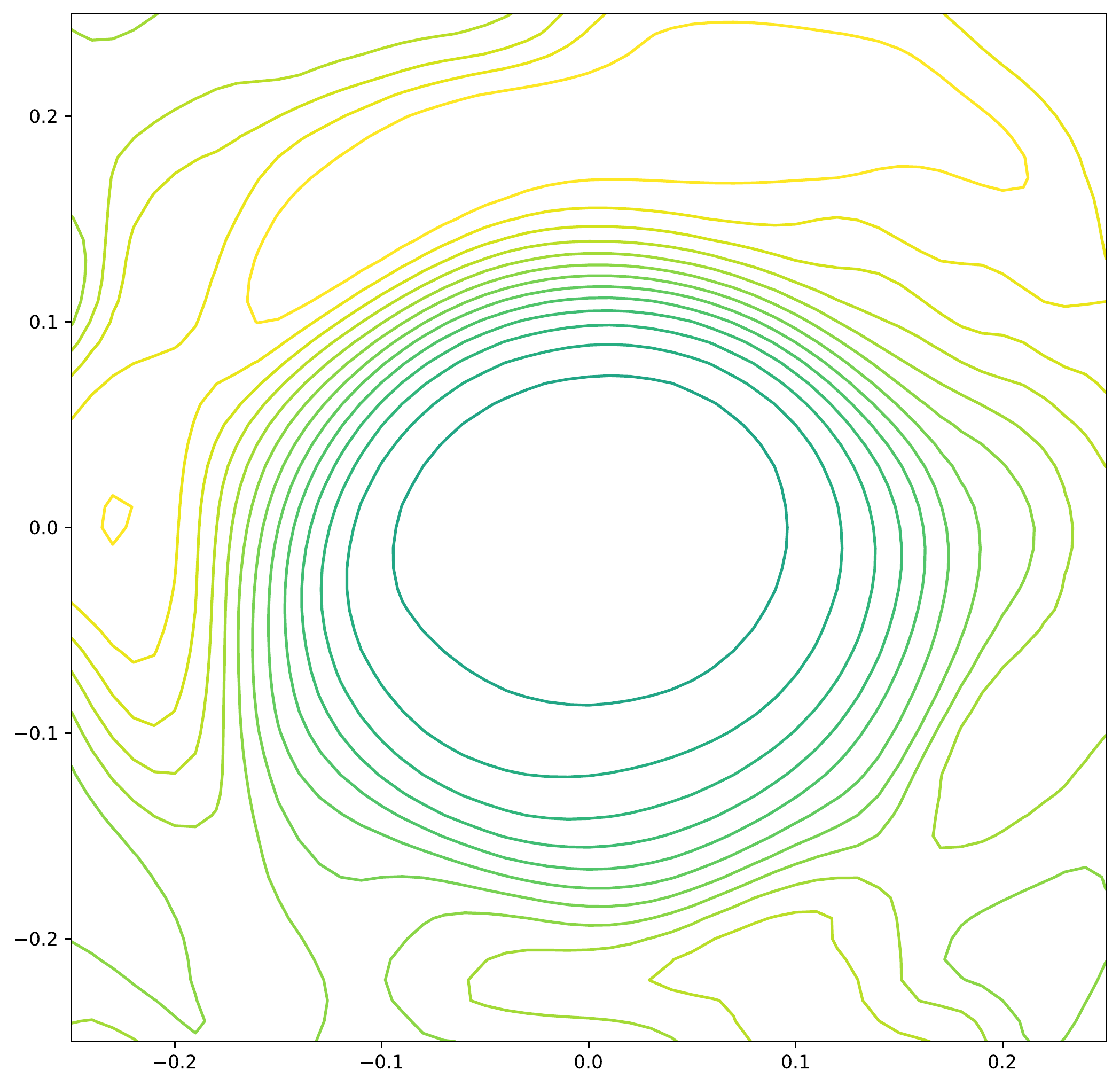}
    	\end{minipage}
% 	\label{fig:ll-2D-first}
    }
    \subfigure[ROSE-Second]{
    	\begin{minipage}[b]{0.22\textwidth}
   		\includegraphics[width=1\textwidth]{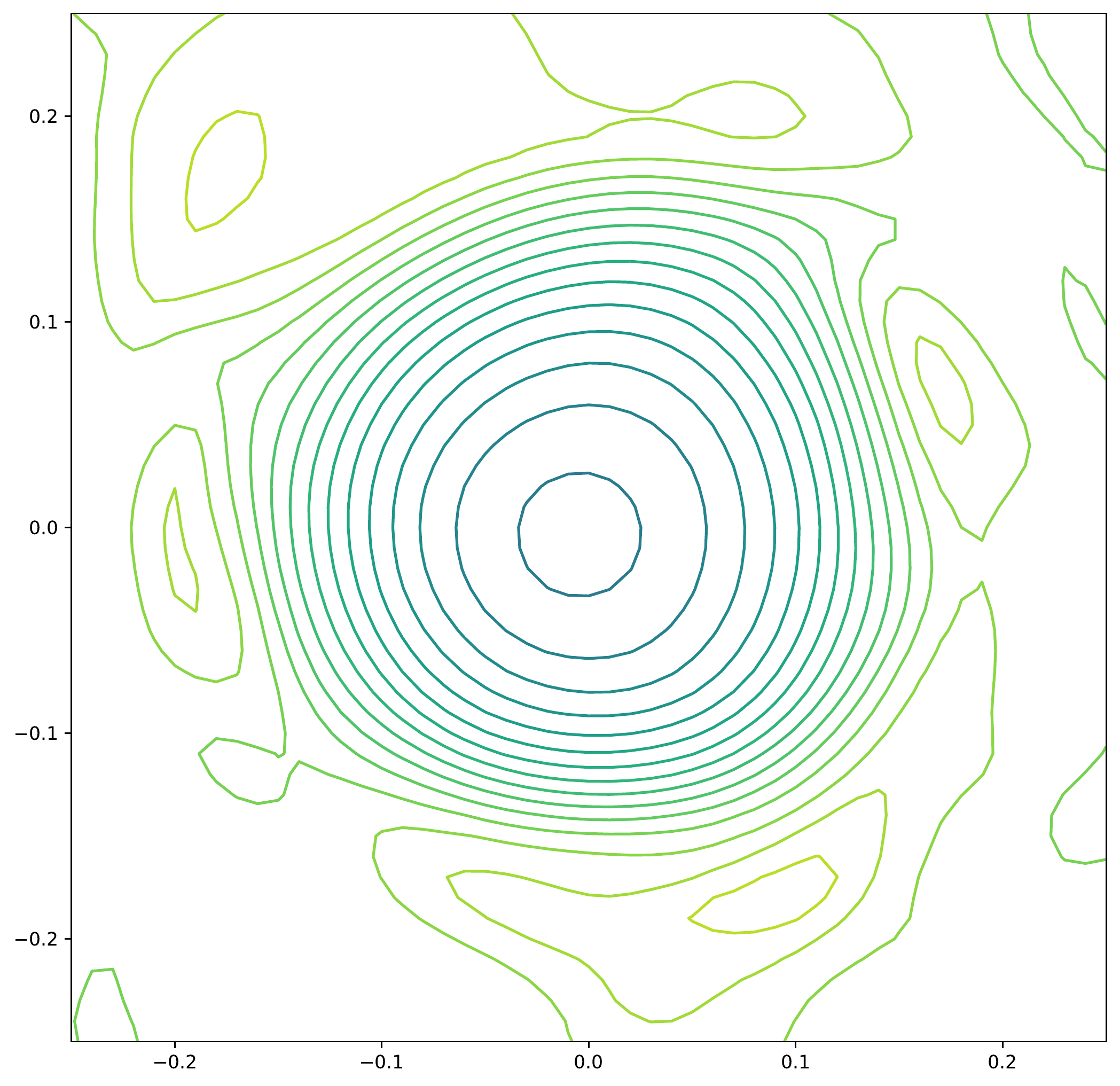}
    	\end{minipage}
% 	\label{fig:ll-2D-second}
    }
    \subfigure[ROSE-Ensemble]{
    	\begin{minipage}[b]{0.22\textwidth}
   		\includegraphics[width=1\textwidth]{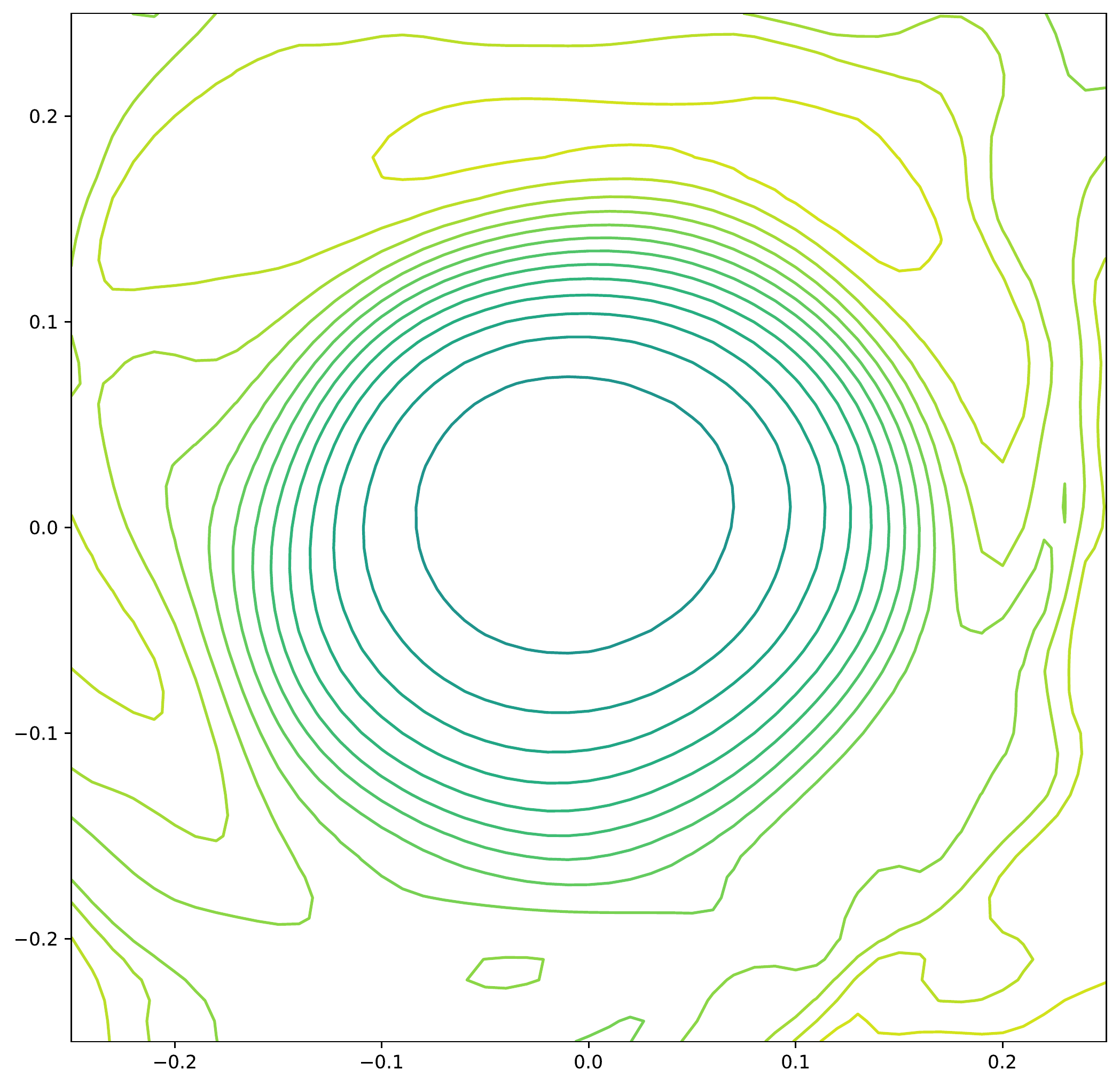}
    	\end{minipage}
% 	\label{fig:ll-2D-ensemble}
    }
    \caption{2D loss contours of models with vanilla fine-tuning and ROSE on four tasks. The lighter color of the contour lines correlates with a larger loss, and the area with denser contour lines demonstrates steeper loss surfaces.}
	\label{fig:2D-ll}
\end{figure*}
Let $\boldsymbol{\theta}$ denote the parameters of a model fine-tuned on downstream tasks.
Then the two-dimensional loss curve of model can be plotted with the function:
\begin{eqnarray}
\label{eq:7}
    f(\alpha, \beta) = \mathcal{L}(\boldsymbol{\theta} + \alpha\boldsymbol{\delta} + \beta\boldsymbol{\eta}),
\end{eqnarray}
where $\mathcal{L}$ is the loss function, and $\alpha, \beta$ are scalar values.
$\boldsymbol{\delta}, \boldsymbol{\eta}$ are direction vectors randomly sampled from Gaussian distribution, which denote two direction vectors in the parameter space corresponding to the two axes of the loss surface.
To remove the scaling effect of neural nets, we follow the filter-wise normalization in \newcite{DBLP:conf/nips/Li0TSG18}, which scales the $\boldsymbol{\delta}, \boldsymbol{\eta}$ to the same norm as parameters by $\frac{\boldsymbol{\delta}}{\Vert\boldsymbol{\delta}\Vert}\Vert\boldsymbol{\theta}\Vert, \frac{\boldsymbol{\eta}}{\Vert\boldsymbol{\eta}\Vert}\Vert\boldsymbol{\theta}\Vert$.
We set the range of both $\alpha$ and $\beta$ to $[-0.25,0.25]$ and uniformly sample $51$ points for each axis.
Since the parameter space is high-dimensional, experimental results confirm the two directions $\boldsymbol{\delta}$ and $\boldsymbol{\eta}$ are divergent and orthogonal to each other.
We plot and compare the loss surfaces of models with vanilla fine-tuning and ROSE on four tasks.

The visualizations are shown in Figure \ref{fig:2D-ll}.
We can observe that ROSE has a significant influence on the smoothness of the loss landscapes across all datasets.
Models fine-tuned with ROSE provide wider and less dense loss contours than vanilla fine-tuning, which shows that they are more robust against noisy perturbations.
Specifically, ROSE-First finds solutions with wider bottoms, and ROSE-Second leads to solutions with less dense loss contour.
This indicates that ROSE-First and ROSE-Second succeed in defense of local and global perturbations, respectively.
Additionally, ROSE-Ensemble is shown to have both of these features, demonstrating that it aggregates the benefits of the two strategies discussed above.
Appendix \ref{sec:ac-1} provides an additional one-dimensional visualization.

\subsection{Probing Preference for Different Features}

We then employ the probing task from \cite{warstadt-etal-2020-learning} to test whether models fine-tuned with ROSE prefer linguistic rather than superficial features.
In the probing experiment, a model is first trained on ambiguous data which equally supports both linguistic and superficial cues, and then tested on disambiguating data which supports only the linguistic cues.
The preference of models for features is measured through Matthews correlation scores between predictions and labels on test sets.
The models are shown a systematic preference for linguistic features if the score is $1$, and complete reliance on superficial cues if the score is $-1$.
Therefore a higher score shows a stronger preference for linguistic features. 
We select two representative experiments gotten by pairing the linguistic feature \textit{Syntactic construction} with two surface features \textit{Lexical content} and \textit{Length}.
For each probing task, we report results of adapted models with different fine-tuning strategies on $5$ random seeds.

\begin{figure}[h]
    \centering
    \subfigure[Lexical content]{
    \begin{minipage}[b]{0.46\linewidth}
      \includegraphics[width=1\textwidth]{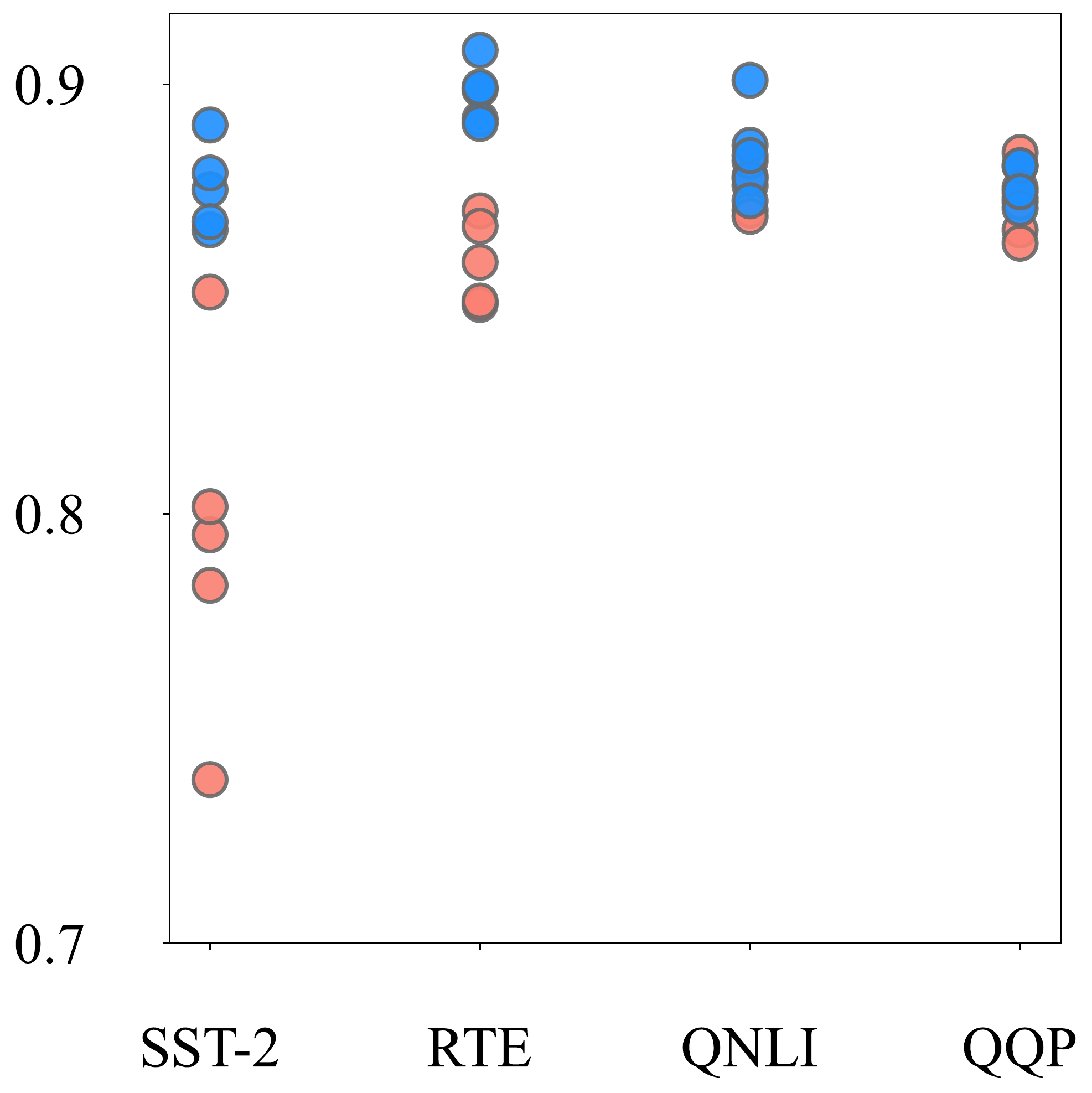}
    \end{minipage}
    }
    \subfigure[Length]{
    \begin{minipage}[b]{0.46\linewidth}
      \includegraphics[width=1\textwidth]{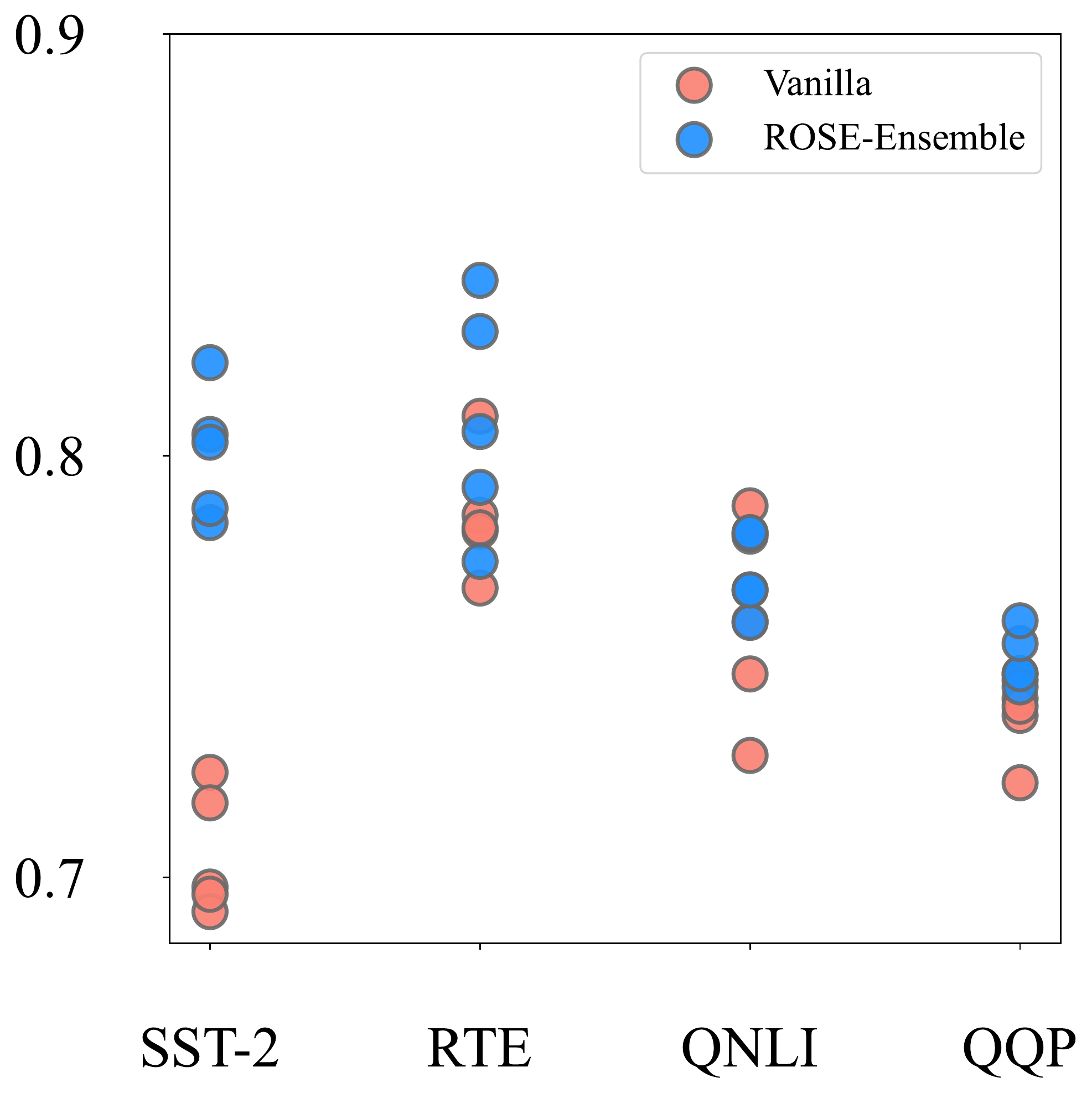}
    \end{minipage}
    }
    \caption{Results for probing tasks with different random seeds.
    Each data point represents one run.
    Models with a stronger preference for linguistic rather than spurious features achieve higher scores.}
    \label{fig:probing}
\end{figure}

Results are plotted in Figure \ref{fig:probing}.
We can observe that, compared to vanilla fine-tuned models, ROSE-tuned models show a stronger preference for linguistic features than any superficial features.
This indicates that ROSE successfully enables models to extract deeper linguistic knowledge during fine-tuning, instead of adopting spurious cues from the training data of downstream tasks.
\section{Related Work}

Adversarial training is the most effective and promising strategy to improve the adversarial robustness of models.
Existing adversarial training methods usually employ PDG-based attacks to generate adversarial examples, and force models to maintain the output on them \cite{DBLP:conf/iclr/ZhuCGSGL20,ijcai2018-0585,wang2021Adversarial}.
Despite the substantial improvements in robustness, adversarial training often requires significant computational and memory costs and fails to preserve the original labels.
Some works focus on constructing reliable adversarial datasets \cite{gardner-etal-2020-evaluating,eger-benz-2020-hero}, which require huge human annotation and only work for a single task.
By contrast, our proposed ROSE is much more efficient and only employs such perturbations to select robust parameters to tune, therefore, there is no need for reliable adversarial examples.

Besides adversarial training methods, our work also relates to a few works of regularization and optimization. 
In regularization, lots of methods have been proposed, including $L^2$-penalty \cite{pmlr-v80-schwarz18a,pmlr-v80-li18a,chen-etal-2020-recall}, weight decay \cite{Kang_Li_Tao_2016,DBLP:conf/iclr/0007WKWA21}, Mixout regularization \cite{DBLP:conf/iclr/LeeCK20}, and so on.
The general approach is augmenting the vanilla optimizer with terms that indirectly or directly penalize the aggressive updates.
Although these methods are exciting, the regularization is often not scalable, and hard to transfer to another model.
Another line of work \cite{wang2021infobert,DBLP:conf/nips/DongLLYZ21} attempts to address this issue from an informative-theoretic perspective.
In optimization, there has been some work proposed recently to force optimization towards wide valleys \cite{DBLP:conf/iclr/ChaudhariCSLBBC17,DBLP:conf/acl/JiangHCLGZ20}.
Compared to these works, ROSE uses the simplest idea by selecting parameters with the second-order robustness in fine-tuning stage to smooth the optimization trajectory.
ROSE is more efficient and can be incorporated into existing methods to improve their adversarial robustness further.

Note that our method does not fall within the realm of model compression.
The target of model compression is to obtain an efficient sub-network with competitive performance, with typical approaches to abandon some parameters when models do inference.
While ROSE aims to improve the adversarial robustness of pre-trained language models, which is done via conducting selective parameters updates in the backward process.

\section{Conclusion}
In this work, we propose an attack-agnostic and model-agnostic defense approach called ROSE, which selectively updates robust parameters during the fine-tuning stage.
We present first-order ROSE which selects the parameters robust against slight perturbation in the hidden space, second-order ROSE which filters out aggressive updates, and ensemble ROSE which aggregates the benefits of the above two strategies.
Experimental results show that both our ROSE-First and ROSE-Second greatly improve the robust performance on various NLP benchmarks, while ROSE-Ensemble is even more effective.
Besides, existing methods achieve better robustness when incorporated with our ROSE. 
We also demonstrate empirically that the effectiveness of ROSE can be attributed to the wider and flatter solutions it finds than the conventional fine-tuning methods.
We hope ROSE could motivate more defense works for language models.

\section*{Limitations}
Although ROSE achieves superior adversarial robustness on four datasets, there are still two limitations.
First, there are some vital hyper-parameters in ROSE, \textit{e.g.} the scaling
coefficient $\gamma$, which have a great influence on the performance as shown in Section \ref{sec:3.6}.
We adopt grid search to select the best parameters, which requires considerable GPU resources.
There is still a need for a more automatic method.
Once we further understand the inner working mechanism of deep neural networks, such hyper-parameters could be calculated theoretically.
Second, due to the limitation of computational resources, we focus on fine-tuning in this work, leaving applying ROSE to pre-training for future work.
We hope ROSE could provide a new perspective for general defense work towards more robust language models.

% Entries for the entire Anthology, followed by custom entries
% \nocite{*}
\bibliography{anthology,custom}
\bibliographystyle{acl_natbib}

\appendix
\section{Appendix}
% The statistics of datasets are shown in Table \ref{tab:statistics}.
% \begin{table}[h]
%     \centering
%     \begin{tabular}{lrrr}
%         \toprule
%         \multirow{2}{*}{\textbf{Dataset}} & \textbf{\# Train} & \textbf{\# Dev} &\textbf{\# Dev} \\
%         & \small GLUE & \small GLUE & \small AdvGLUE \\
%         \hline \hline
%         SST-2 & 67k & 0.9k & 149 \\
%         RTE & 2.5k & 0.3k & 82 \\
%         QNLI & 108k & 5.5k & 149 \\
%         QQP & 364k & 40k & 79 \\
%         \bottomrule
%     \end{tabular}
%     \caption{Statistics of the datasets we used. All tasks are binary classifications and report accuracy.}
%     \label{tab:statistics}
% \end{table}

\subsection{One-dimensional Linear Interpolation}
\label{sec:ac-1}
In order to investigate the robustness of our method, we present parametric one-dimensional visualization as described in \cite{DBLP:journals/corr/GoodfellowV14}, which plots the value of loss function along the line connecting two different models.
Let $\boldsymbol{\theta}$ and $\boldsymbol{\theta}^{'}$ indicate the parameters of these two models respectively.
Then we plot the function:
\begin{eqnarray}
\label{eq:6}
    f(\alpha) = \mathcal{L}((1-\alpha)\boldsymbol{\theta} + \alpha\boldsymbol{\theta}^{'}),
\end{eqnarray}
where $\alpha$ is a scalar value.
We compare the weights of models obtained by vanilla and ROSE-Ensemble fine-tuning method on four tasks.
In particular, for $\alpha\in[-0.5, 1.5]$, we uniformly sample $51$ points and plot the function $f(\alpha)$ and superimpose the classification accuracy.

\begin{figure}[h]
	\centering
	\subfigure[SST-2]{
		\begin{minipage}[b]{0.46\linewidth}
			\includegraphics[width=1\textwidth]{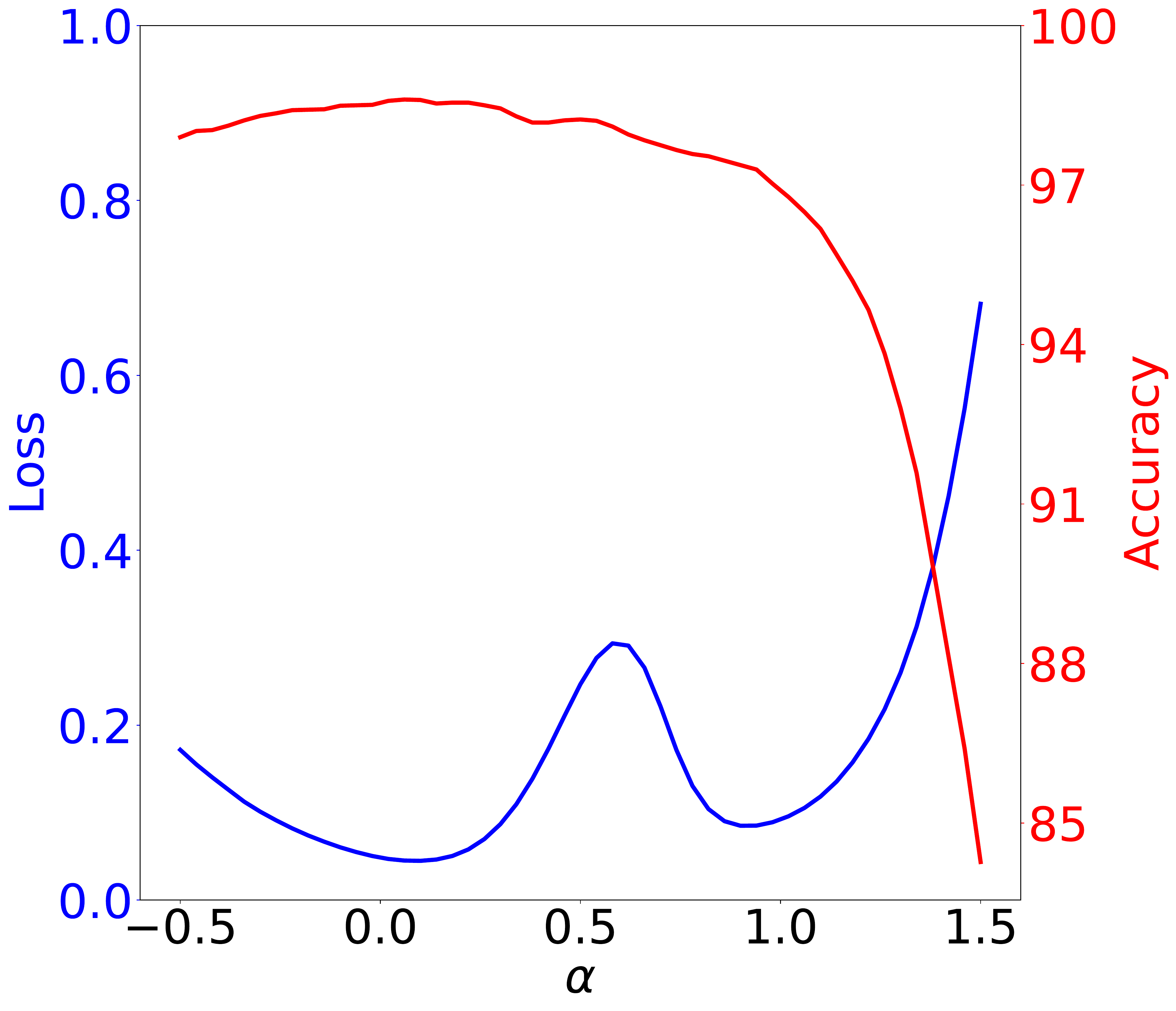}
		\end{minipage}
% 		\label{fig:ll-1D-vanilla}
	}
    \subfigure[RTE]{
    	\begin{minipage}[b]{0.46\linewidth}
   		\includegraphics[width=1\textwidth]{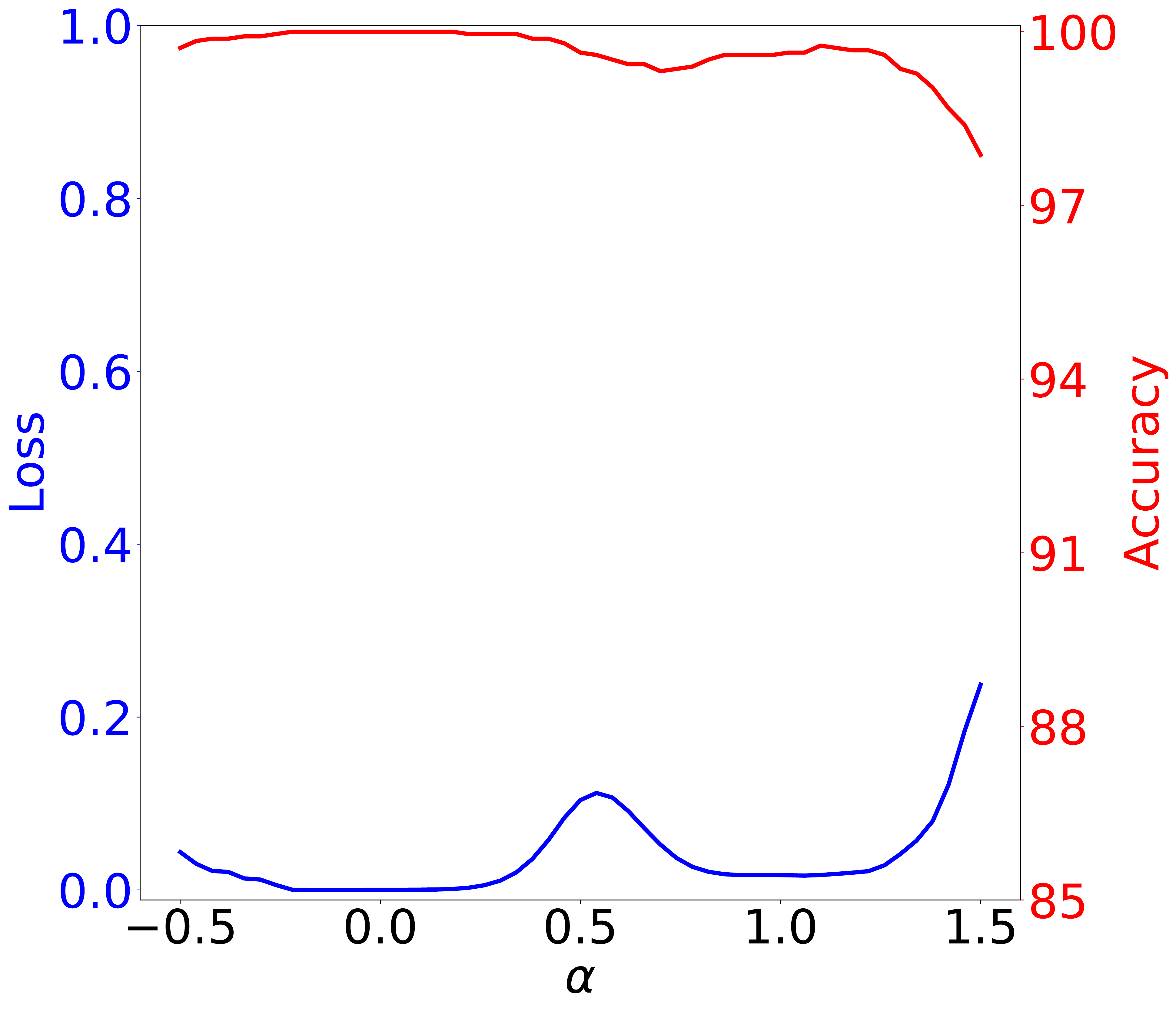}
    	\end{minipage}
    }
    \\
    \subfigure[QNLI]{
    	\begin{minipage}[b]{0.46\linewidth}
   		\includegraphics[width=1\textwidth]{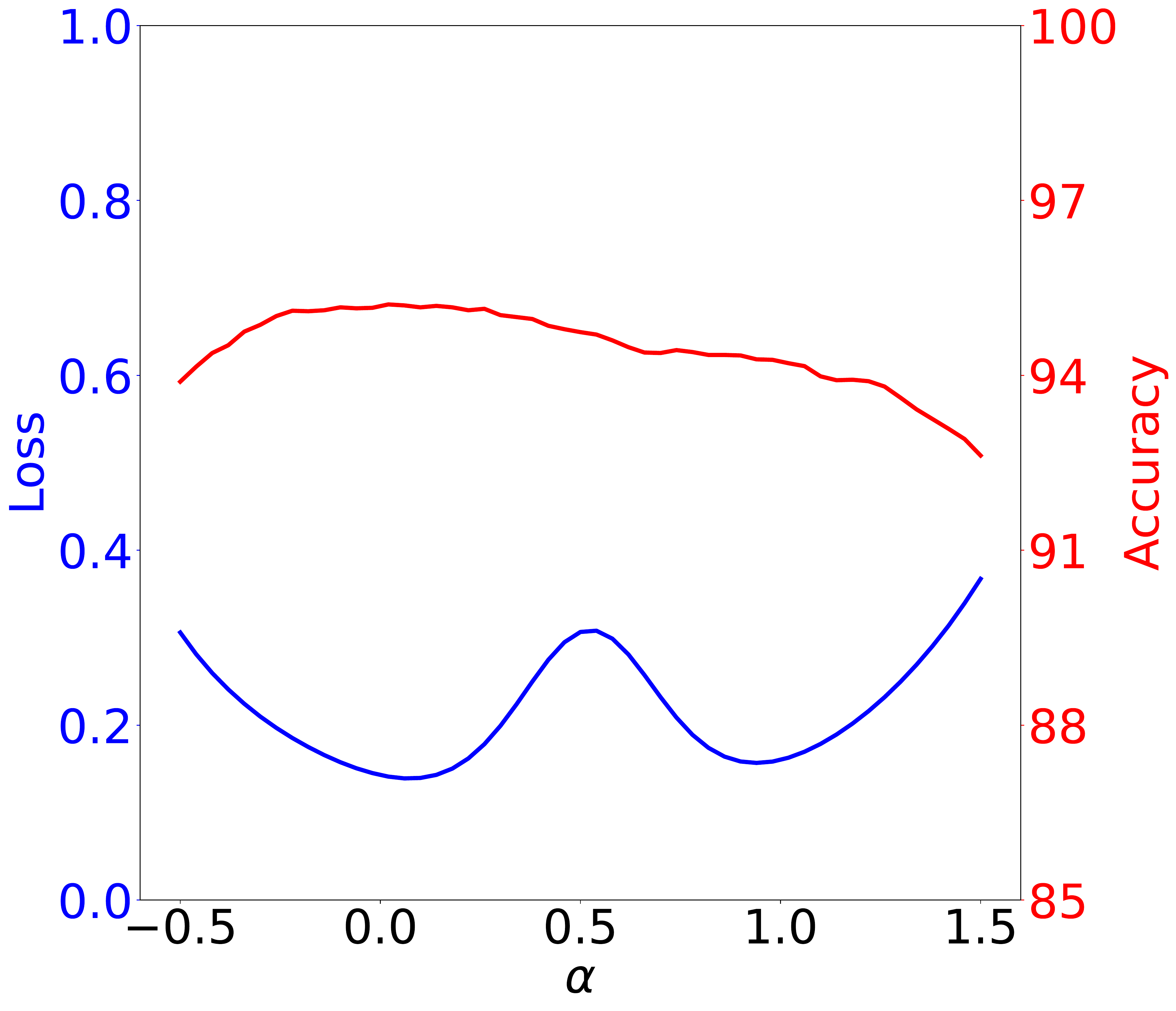}
    	\end{minipage}
    }
    \subfigure[QQP]{
    	\begin{minipage}[b]{0.46\linewidth}
   		\includegraphics[width=1\textwidth]{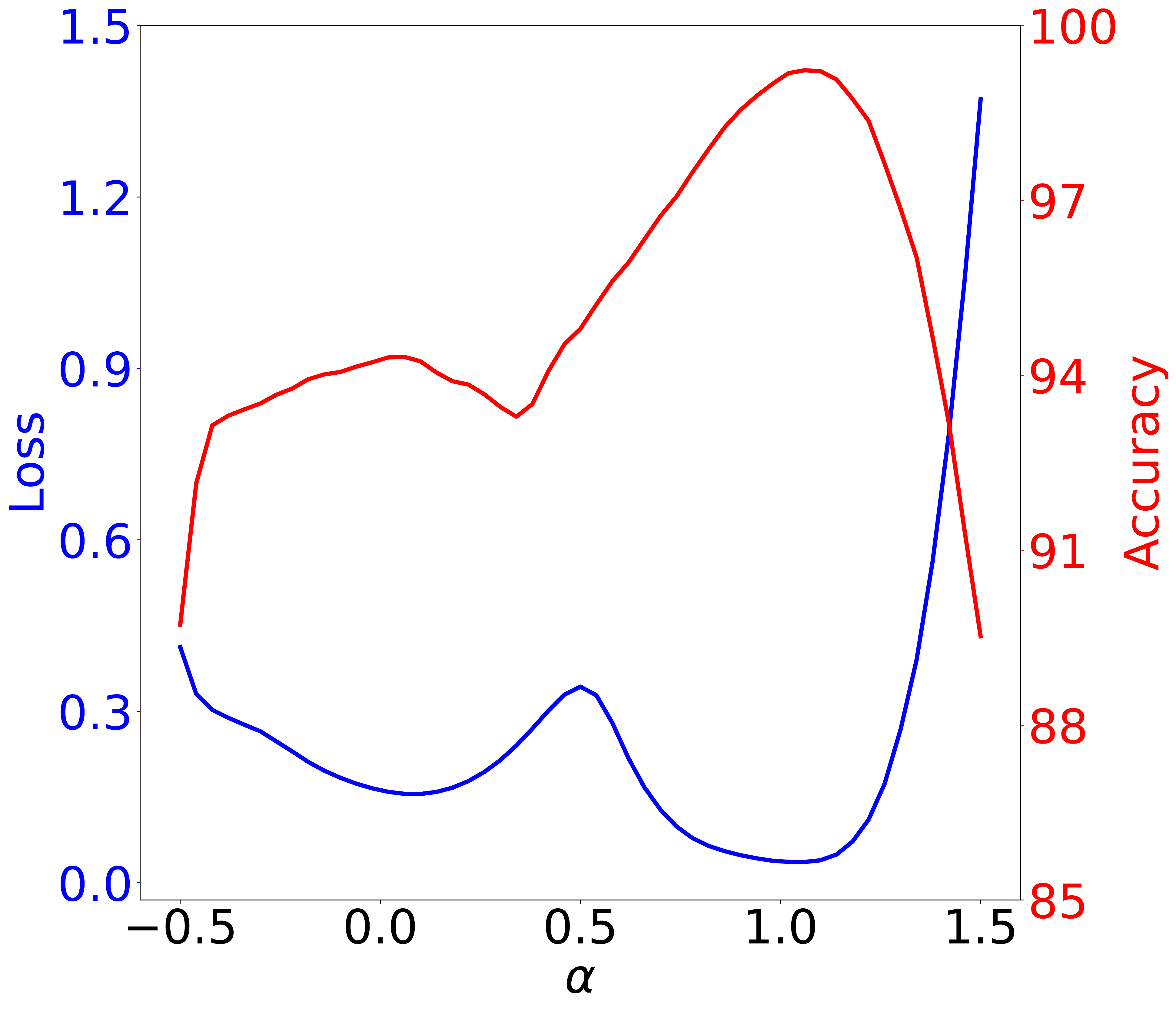}
    	\end{minipage}
    }
	\caption{1D loss interpolation (Left axis corresponds to loss, and right to accuracy) between solutions found by vanilla fine-tuning and ROSE-Ensemble on four tasks. Models trained with ROSE ($\alpha=0$) have flatter and wider curves than vanilla fine-tuning ($\alpha=1$), which correlates well with robustness.}
	\label{fig:1D-ll}
\end{figure}

Figure \ref{fig:1D-ll} shows the visualization of fine-tuned models with different strategies.
Compared with vanilla fine-tuning, we can observe that our ROSE provides wider and flatter curves.
The result indicates that solutions obtained by ROSE tend to be more robust.

\subsection{Effectiveness of Dropout}
In order to investigate the effectiveness of dropout in ROSE-First, For two outputs produced by different dropout, we inspect the ratio (\%) of prediction labels that are not consistent with each other on the first 200, 600, 2000 and 4000 steps. 

\begin{table}[htbp]
    \centering
    \begin{tabular}{l|c|c|c|c}
    \toprule
    Step & 200 & 600 & 2000 & 4000 \\
    \hline
    SST-2 &	12.69 & 7.80 &	5.25 & 4.45 \\
    RTE	& 13.30 & 9.44 & 5.12 & - \\
    QNLI & 18.44 & 11.61 & 7.70 & 6.32 \\
    QQP & 12.31 & 9.58 & 7.35 & 6.45 \\
    \bottomrule
    \end{tabular}
    \caption{Results for ROSE-Ensemble with different $\gamma$.}
    \label{tab:dropout-effect}
\end{table}

From Table \ref{tab:dropout-effect} we can see that, dropout generates adversarial examples as expected with striking low cost. The ratio decreases fast at the beginning and comes to stable finally, which indicates that ROSE-First succeeds to improve the robustness of models against such perturbation over the training process.

\end{document}